\documentclass{article}

\usepackage{PRIMEarxiv}

\usepackage[utf8]{inputenc} 

\usepackage{amsmath,amsfonts}
\usepackage{algorithmic}
\usepackage{algorithm}
\usepackage{array}
\usepackage[caption=false,font=normalsize,labelfont=sf,textfont=sf]{subfig}
\usepackage{textcomp}
\usepackage{stfloats}
\usepackage{url}
\usepackage{verbatim}
\usepackage{graphicx}
\usepackage{booktabs}
\usepackage{makecell} 
\usepackage{soul, color}
\usepackage{flushend}
\usepackage{ulem}
\usepackage{enumitem}

\usepackage{lineno,hyperref}
\modulolinenumbers[5]
\hypersetup{colorlinks = true,linkcolor = blue,anchorcolor =red,citecolor = blue,filecolor = red,urlcolor = red,
            pdfauthor=author}

\title{TFN: An Interpretable Neural Network with Time-Frequency Transform Embedded for Intelligent Fault Diagnosis
}

\author{
  Qian Chen \\
  Shanghai Jiao Tong University \\
  \texttt{chenqian2020@sjtu.edu.cn} \\
     \And
  Xingjian Dong \thanks{Corresponding Author.} \\
  Shanghai Jiao Tong University \\
  \texttt{donxij@sjtu.edu.cn} \\
     \And
  Guowei Tu\\
  Shanghai Jiao Tong University \\
  \texttt{guoweitu@sjtu.edu.cn} \\   
  \And
  Dong Wang\\
  Shanghai Jiao Tong University \\
  \texttt{dongwang4-c@sjtu.edu.cn} \\   
    \And
  Baoxuan Zhao\\
  Shanghai Jiao Tong University \\
  \texttt{bxzhao@sjtu.edu.cn} \\
     \And
  Zhike Peng\\
  Shanghai Jiao Tong University \\
  \texttt{z.peng@sjtu.edu.cn} \\
}

\begin{document}
\maketitle

\begin{abstract}
Convolutional Neural Networks (CNNs) are widely used in fault diagnosis of mechanical systems due to their powerful feature extraction and classification capabilities. However, the CNN is a typical black-box model, and the mechanism of CNN's decision-making is not clear, which limits its application in high-reliability-required fault diagnosis scenarios. To tackle this issue,  we propose a novel interpretable neural network termed as Time-Frequency Network (TFN), where the physically meaningful time-frequency transform (TFT) method is embedded into the traditional convolutional layer as a trainable preprocessing layer. This preprocessing layer named as time-frequency convolutional (TFconv) layer, is constrained by a well-designed kernel function to extract fault-related time-frequency information. It not only improves the diagnostic performance but also reveals the logical foundation of the CNN prediction in a frequency domain view. Different TFT methods correspond to different kernel functions of the TFconv layer. In this study, three typical TFT methods are considered to formulate the TFNs and their diagnostic effectiveness and interpretability are proved through three mechanical fault diagnosis experiments.  Experimental results also show that the proposed TFconv layer has outstanding advantages in convergence speed and few-shot scenarios, and can be easily generalized to other CNNs with different depths to improve their diagnostic performances.  The code of TFN is available on \textit{\url{https://github.com/ChenQian0618/TFN}}.
\end{abstract}


\section{Introduction}

Nowadays, fault diagnosis of mechanical equipment is widely used in industry to reduce property damage and improve production efficiency \cite{liRollingBearingFault2021,LI2015330}. With the development of sensor technology and the industrial internet, a large amount of operation and maintenance data could be obtained by various sensors \cite{taoDatadrivenSmartManufacturing2018}, such as accelerometers, dynamometers, and microphones. Based on sufficient operation and maintenance data, the data-driven method \cite{chenDataDrivenFaultDiagnosis2022}, as a model-free solution with high diagnostic accuracy, has gradually gained more and more attention in the field of mechanical equipment fault diagnosis. 

The process of data-driven fault diagnosis can be divided into three steps: data acquisition, feature extraction, and fault classification \cite{leiApplicationsMachineLearning2020}. Among these steps, feature extraction is the key to fault diagnosis \cite{liRotationalMachineHealth2012, siKeyPerformanceIndicatorRelatedProcessMonitoring2021}. Traditional feature extraction mainly relies on signal processing methods, including Fourier Transform (FT), Short Time Fourier Transform (STFT), Wavelet Transform (WT) \cite{pengApplicationWaveletTransform2004}, and Empirical Mode Decomposition (EMD) \cite{liApplicationBandwidthEMD2017}. Using these methods, researchers can transform raw maintenance data (i.e., vibration signals) from the time domain into the frequency or time-frequency domain, and thus extract essential features for fault classification. However, such a strategy requires too much expertise and prior knowledge, which is difficult to be  widely conducted in real industrial scenarios. Therefore, deep learning, conducting feature extraction and classification automatically, becomes the promising solution for mechanical fault diagnosis \cite{leiApplicationsMachineLearning2020}.

As a powerful representation learning technique, deep learning provides end-to-end solutions and is widely used in computer vision\cite{brunettiComputerVisionDeep2018}, natural language processing\cite{fayekEvaluatingDeepLearning2017}, game competition\cite{silverMasteringGameGo2017}, and other fields. In terms of fault diagnosis, deep learning has a series of nonlinear mapping layers to extract hidden key information from vibration signals. Many deep learning models have been applied to mechanical fault diagnosis, such as deep belief network (DBN) \cite{shaoElectricLocomotiveBearing2018}, convolutional neural network (CNN) \cite{wangCoarsetoFineProgressiveKnowledge2021,zhaoIntelligentFaultDiagnosis2022,pengMultibranchMultiscaleCNN2020}, and recurrent neural network (RNN) \cite{nieNovelNormalizedRecurrent2021}. Among these neural networks, CNN-based models can fully extract the spatial information of the input signal and thus leads to high diagnostic accuracy in multiple public datasets for mechanical fault diagnosis \cite{zhaoDeepLearningAlgorithms2020}.

Despite its superior diagnostic performance, the CNN has a weak spot - its interpretability \cite{zhangVisualInterpretabilityDeep2018}. It is difficult to find the logical foundation of the CNN model for feature extraction and classification. This reduces the credibility of the results and prohibits the breakthrough of the model performance, which in turn limits its application in high-reliability-required fault diagnosis scenarios (e.g. aero engine fault diagnosis \cite{xiLeastSquaresSupport2019}). Therefore, it is of great importance to develop interpretable CNNs for high-quality fault diagnosis of mechanical systems.

The research of interpretable CNN is a growing topic in the AI field \cite{ivanovsPerturbationbasedMethodsExplaining2021,fanInterpretabilityArtificialNeural2021}. Current interpreting methods can be roughly divided into four categories \cite{zhangSurveyNeuralNetwork2021}: rule type, semantic type, attribution type, and example type. Rule type is to extract the mapping relationship of CNNs into specific logical rules (e.g. CEM\cite{dhurandharExplanationsBasedMissing2018}, CDRP\cite{wangInterpretNeuralNetworks2018}). Semantic type is to completely analyze the meaning of specific hidden layers in CNNs (e.g. Network dissection\cite{bauNetworkDissectionQuantifying2017}). Attribution type is to quantify the contribution (or negative effect) of the input and features (e.g. salient map \cite{simonyanDeepConvolutionalNetworks2014}, Grad-CAM\cite{selvarajuGradCAMVisualExplanations2017}, IG \cite{sundararajanAxiomaticAttributionDeep2017}). Example type is to summarize typical samples related to the category (e.g. Influence functions\cite{kohUnderstandingBlackboxPredictions2017}). The above interpreting methods are mainly designed for 2D images, and when it comes to 1D vibration signals in mechanical fault diagnosis, these techniques are not suitable.

At present, there are few studies to explore the interpretability of CNNs in the field of mechanical fault diagnosis. Li \textit{et al}.\cite{liWhiteningNetGeneralizedNetwork2021} calculated the importance weight of each point of the vibration signal sample to the CNN prediction through the integrated gradient (IG) method, and they explained the foundation of prediction-making through the frequency spectra of calculated weights. Wu \textit{et al}. \cite{wuHybridClassificationAutoencoder2021} transformed 1D vibration signals into 2D time-frequency spectra as the input samples, so that the 2D interpreting method (i.e., Grad-CAM) could be adopted to obtain the model attention area. Wang \textit{et al}. \cite{wangFullyInterpretableNeural2022} extended the extreme learning machine (ELM) to an interpretable structure for machine state monitoring. With this strategy, the information frequency band can be automatically located. Zhao \textit{et al}. \cite{zhaoInterpretableDenoisingLayer2021} incorporated the Reproducing Kernel Hilbert Space (RKHS) into the convolutional layer and built a specific interpretable denoising convolutional layer. Li \textit{et al}. \cite{liWaveletKernelNetInterpretableDeep2022} combined continuous wavelet transform with convolutional layers to formulate a wavelet kernel network to extract interpretable features. The above studies help to explain the mechanism of CNNs in fault diagnosis to some extent, but their interpretations are usually ambiguous and need subjective understandings (e.g. the similarity between different frequency spectra of inputs \cite{liWhiteningNetGeneralizedNetwork2021}, the sensibility of CNN to impulsive signal \cite{liWaveletKernelNetInterpretableDeep2022}). On top of that, these strategies may bring some other issues: degraded diagnostic performance and poor generalizability.

To propose an interpretable CNN model for fault diagnosis, we set our sights on traditional signal processing methods which are physically meaningful and good at feature extraction. Considering that the Fourier-type transform and the convolutional layer can be both regarded as inner products, we embed the time-frequency transform (TFT) method into the traditional convolutional layer. This leads to a novel layer named time-frequency convolutional (TFconv) layer, which is constrained by a well-designed kernel function in order to extract fault-related time-frequency information. Using the TFconv layer as the preprocessing layer, we constructed the Time-Frequency Network (TFN) for fault diagnosis tasks. With this interpretable TFN, we not only improve the accuracy of fault diagnosis, but also reveal the logical foundation of prediction-making in the frequency domain through the frequency response analysis. A series of mechanical fault diagnosis experiments verify the superior diagnostic performance and the clear interpretability of TFN. 

The idea of combining signal processing methods with neural networks has already been proposed in the previous works as shown in Table \ref{tab:comparison with our work}, but our method has enough novelty and is distinguished in the following three aspects. 1) Distinctive motivation: Our method tries to parameterize the convolutional layer to simulate time-frequency transform which is very novel against the existing literature;  2) Considering complex value kernel: Previous works\cite{liWaveletKernelNetInterpretableDeep2022,ravanelliInterpretableConvolutionalFilters2019a,gangulyWaveletKernelBased2020} are trying to initialize or parameterize real value kernel and are equivalent to FIR filter, whose output is the filtered sub-signals, while our work takes complex value kernel in consideration and is equivalent to time-frequency transform, whose output is the energy distribution of the signal in the time-frequency domain.  3) A win-win of performance and interpretability: our proposed method not only improves the accuracy of fault diagnosis, but also explains the focusing frequency of CNN models.
\begin{table}[htbp]
\caption{The representative works related to our method.\label{tab:comparison with our work}}
\centering
\begin{tabular}{cll}
\toprule[1pt]
\textbf{Method}& \textbf{Application} & \textbf{Motivation}\\
\midrule[0.5pt]
\vspace{8pt}
SincNet, 2019\cite{ravanelliInterpretableConvolutionalFilters2019a} & speaker recognition & \parbox[c]{8.5cm}{\textbf{parameterize} \textbf{real} value convolutional kernel by \textbf{trainable sine} function}\\
\vspace{8pt}
W-CNN, 2020\cite{gangulyWaveletKernelBased2020} & discharge detection & \parbox[c]{8.5cm}{\textbf{initialize} \textbf{real} value convolutional kernel by \textbf{wavelet} function}\\
\vspace{8pt}
WKN, 2022\cite{liWaveletKernelNetInterpretableDeep2022} & fault diagnosis & \parbox[c]{8.5cm}{\textbf{parameterize} \textbf{real} value convolutional kernel by \textbf{trainable wavelet} function}\\
\vspace{8pt}
DeSpaWN, 2022\cite{michauFullyLearnableDeep2022} & unsupervised monitoring & \parbox[c]{8.5cm}{design a new layer to simulate \textbf{wavelet decomposition}}\\
\parbox[c]{2.5cm}{\centering TFN\\ (our method)} & fault diagnosis & \parbox[c]{8.5cm}{  \textbf{parameterize} \textbf{complex} value convolutional kernel to simulate \textbf{time-frequency transform} and \textbf{interpret the attention that CNN paid to different frequencies}}\\
\bottomrule[1pt]
\end{tabular}
\end{table}

The main contributions of this study could be summarized as follows:
\begin{enumerate}
        \item TFconv layer with excellent interpretability is proposed, with which we can extract time-frequency information to obtain a better diagnostic performance.
        \item All the inner product based TFT methods can be embedded into the TFconv layer as kernel functions, and three typical TFT methods are considered to formulate corresponding TFconv layers and further be analyzed. 
        \item TFN is proposed by combining the interpretable TFconv layer with a backbone CNN, and we depict the entire procedure of using the TFN in mechanical fault diagnosis.
        \item Frequency response analysis is performed on the well-trained TFconv layer to explain the logical foundation of feature extraction and prediction-making of TFN in the frequency domain.
        \item  The proposed TFconv layer has outstanding advantages in convergence speed and few-shot scenarios, and can be easily generalized to other CNN models with different depths to improve their diagnostic performances. 
\end{enumerate} 

The rest of this article is organized as follows. Section \uppercase\expandafter{\romannumeral 2} introduces TFT and CNN as the theoretical foundation. Based on Section \uppercase\expandafter{\romannumeral 2}, the construction of the TFconv layer, the interpreting method, and the diagnosis procedure using TFN are presented in Section \uppercase\expandafter{\romannumeral 3}. In Section \uppercase\expandafter{\romannumeral 4}, three mechanical dataset experiments are carried out to verify the diagnostic performance and interpretability of TFN.  Section \uppercase\expandafter{\romannumeral 5} introduces the essential differences between TFNs and contrast models, and conducts further analyses of training time, convergence speed, few-shot ability and generalizability of proposed TFNs.  Finally, Conclusions are given in Section \uppercase\expandafter{\romannumeral 6}.

\section{Prilimilary}

\subsection{Time-Frequency Transform}

Time-frequency transform (TFT) is widely used in signal processing and it is based on the inner product. Mathematically, the purpose of the inner product is to measure the similarity, and any vector could be decomposed into a set of orthogonal bases by means of the inner product. Considering the 1D vibration signal as a vector, we can decompose the signal in the same way. Fourier transform, the most fundamental signal processing method, takes the frequency-orthogonal sine function as the basis and decomposes the vibration signal through the inner product to obtain the amplitude of the vibration signal at different frequencies, i.e., the frequency spectrum:
\begin{equation}
X(f) = \int_{ - \infty }^{ + \infty } {x(t){{({e^{j2\pi ft}})}^*}dt}  = \left\langle {x(t),{e^{j2\pi ft}}} \right\rangle
\label{eq:FT}
\end{equation}
where $X(f)$ denotes the frequency spectrum, $x(t)$ denotes the analysed signal, $e^{j2\pi ft}$ denotes the trigonometric bases and $*$ denotes the conjugate operator.

Although the frequency spectrum obtained by the inner product with frequency-orthogonal bases can effectively reveal the frequency domain information of the vibration signal, it leaves out the time domain information and is not suitable for non-stationary signals. To process non-stationary signals, it is necessary to obtain the amplitude of the signal at different times and different frequencies, i.e., the time-frequency spectrum.

The time-frequency spectrum can be obtained by the inner product with time-frequency-orthogonal bases, which is formulated as
\begin{equation}
TF(\tau ,f) = \int_{ - \infty }^{ + \infty } {x(t){\psi _f}^*(t - \tau )dt}  = \left\langle {x(t),{\psi _f}(t - \tau )} \right\rangle
\label{eq:TFT}
\end{equation}
where $TF(\tau ,f)$ denotes the time-frequency spectrum, $X(t)$ denotes the analysed signal, $\left\langle {{\psi _f}(t - \tau )}\right\rangle_{f,t} $ denotes the time-frequency-orthogonal bases.  ${\psi _f}(t)$ denotes the inner product window function that has compact support in both time domain and frequency domain.

\begin{figure}[htbp]
        \centering
        \includegraphics[width=8.5 cm]{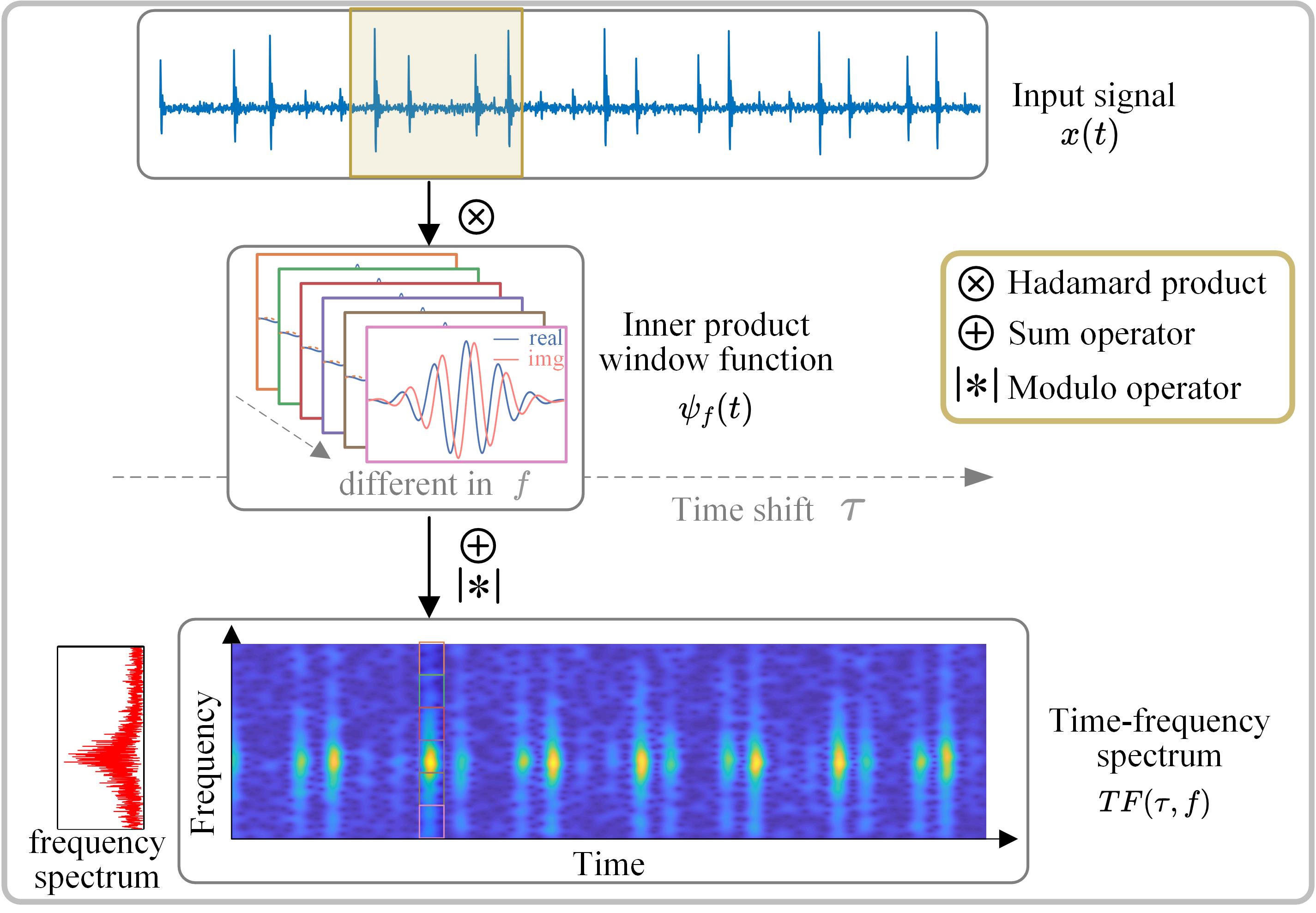}
        \caption{The process of inner product based TFT.}
        \label{fig:TFT process}
\end{figure}

The process of inner product based TFT is demonstrated in Fig. \ref{fig:TFT process}. The input signal $x(t)$ and the inner product window functions $\psi_f(t)$ with different frequency bands are convolved to obtain the frequency spectrum of the signal at a specific time point $\tau$. With the movement of the time point $\tau$, the complete time-frequency spectrum $TF(\tau ,f)$ is gradually obtained. Among the process, parameter $\tau$ and $f$ are used to adjust the focusing area of inner product window function in time domain and frequency domain respectively. It should be noted that the inner product window function is a complex function, so the obtained time-frequency spectrum is also complex, that is, the spectrum includes both the power information and the phase information. In order to separate the power information, the complex modulo operation of the time-frequency spectrum is required.

TFT methods include short-time Fourier transform (STFT), Chirplet Transform (CT)\cite{yangMulticomponentSignalAnalysis2013, tuIterativeNonlinearChirp2020}, and wavelet transform (WT)\cite{cohenBetterWayDefine2019}, etc.  All these methods follow the above procedure and the only difference between them is the inner product window function. STFT constructs an inner product window function by multiplying the sine function by a time domain window. CT introduces a linear frequency modulation factor to STFT, thereby improving the the accuracy and stability of the result under variable speed conditions. WT uses the wavelet family, which is generated by the mother wavelet through scaling and translation, as the inner product window function, to obtain a time-frequency window that varies with frequency. WT has low time domain resolution and high frequency domain resolution at the low frequency, and vice versa at the high frequency. The inner product window functions of the above four TFT methods are shown in Table \ref{tab:kernel Function}.

TFT methods can project the non-stationary signals from time domain into time-frequency domain to obtain their time-frequency joint information.  As shown in Fig. \ref{fig:TFT process}, the input is a simulated signal of rolling bearing with inner fault, and its time-frequency spectrum demonstrates both the vibration frequency and the dual-impulse property. Due to their powerful analytical abilities, TFT methods play an important role in the feature extraction of mechanical fault diagnosis.

\subsection{CNN}
The structure of a convolutional neural network (CNN) can be roughly divided into two parts: the convolutional part and the classification part. The convolutional part consists of a series of convolutional layers, BN layers, activation layers, and pooling layers, while the classification part consists of several fully connected layers. The input samples usually are 1D vibration signals, so the CNN models used for mechanical system fault diagnosis are 1D accordingly.

\begin{figure}[htbp]
        \centering
        \includegraphics[width=8.5 cm]{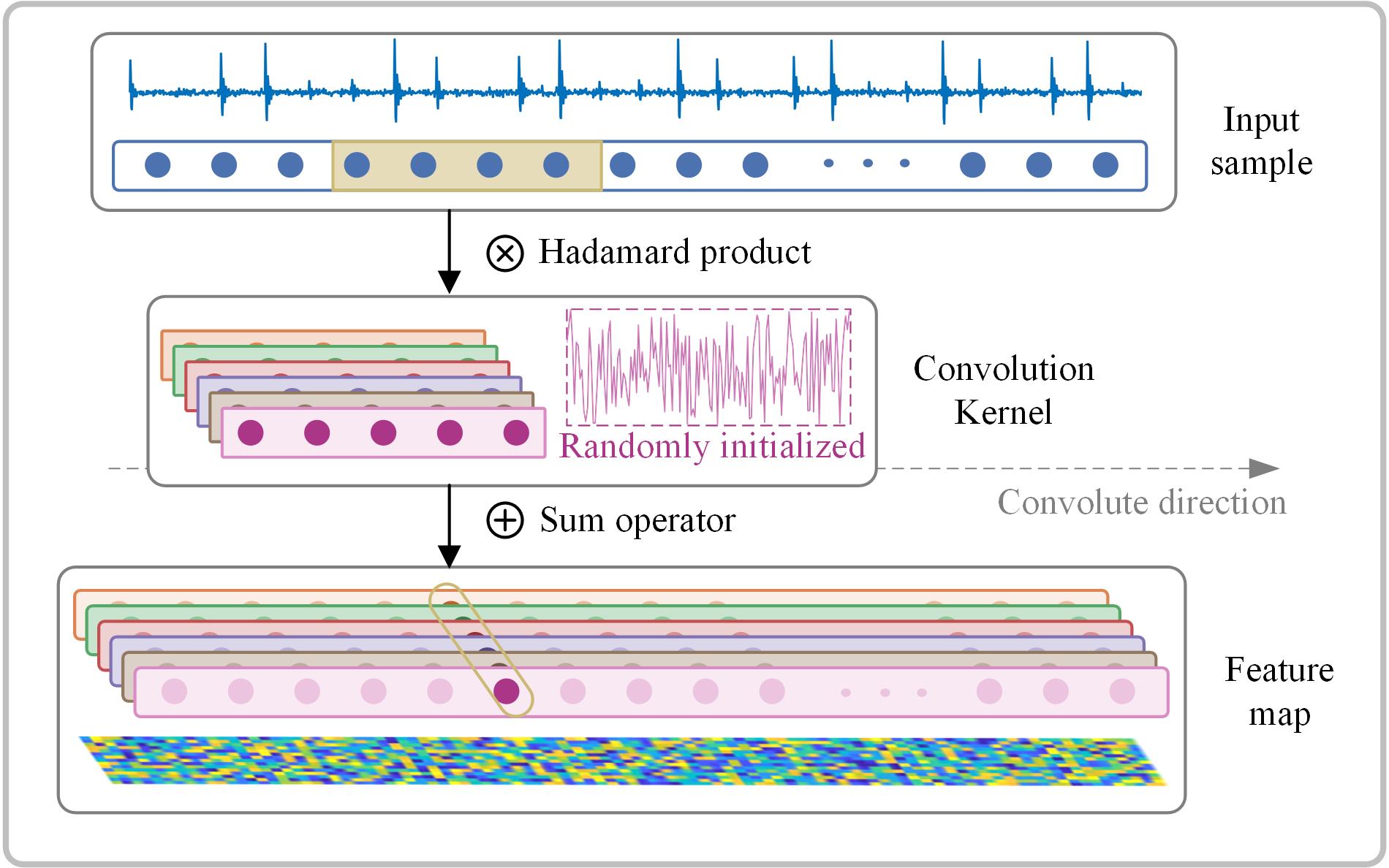}
        \caption{The process of traditional convolutional layer in CNN.}
        \label{fig:CNN process}
\end{figure}

The convolutional layer is the core part of the CNN and the process of 1D convolution is shown in Fig. \ref{fig:CNN process}. Each randomly initialized convolution kernel is convolved along the 1D input signal, and the results of multiple convolution kernels are concatenated to obtain the feature map. The feature map of the $l$-th layer on the $k$-th channel can be expressed as
\begin{equation}
h_k^l = w_k^l*{x^l} + b_k^l
\label{eq:conv}
\end{equation}
where, $x^l$ denotes the input of the $l$-th convolutional layer, and $w_k^l$ and $b_k^l$ denote the weight and bias of the $k$-th convolution kernel in the $l$-th convolutional layer respectively. The symbol $*$ denotes the convolution operator.

The BN layer normalizes inputs to a specific distribution, thereby increasing the training speed and alleviating gradient explosion or disappearance. The output value of the $k$-th channel in the $l$-th BN layer can be given as
\begin{equation}
{\hat h_k^l} = \frac{{{h_k^l} - {\mu_k^l}}}{{\sqrt {{\sigma_k^l}^2 + \varepsilon } }}
\label{eq:BN}
\end{equation}
where $\mu_k^l$ and $\sigma_k^l$ denote the channel-wise mean and variance of the input $h_k^l$, and $\varepsilon$ denotes a small constant that prevents the denominator from being zero.

The activation layer introduces nonlinearity into CNN, thereby enabling nonlinear mapping. The output of the $l$-th activation layer can be denoted as
\begin{equation}
{z^l} = f\left( {{{\hat h}^l}} \right)
\label{eq:activation}
\end{equation}
where ${\hat h}^l$ denotes the input of the activation layer and $f$ denotes the nonlinear activation function, e.g. ReLU, Sigmoid.

With quite many features extracted, the pooling layer aims to compress them to reduce the model parameters while preserve the main information. The output of the $l$-th pooling layer can be denoted as
\begin{equation}
{y^l} = {\rm down} \left( {{z^l}} \right)
\label{eq:pooling}
\end{equation}
where $y^l$ denotes the input and ${\rm down}$ denotes downsampling operation, such as maximum or average downsampling.

After the convolutional part, the raw signal has been transformed into high-dimensional features, which are flattened and further classified in the following classification part. Finally, the {\textit{Softmax}} function is used to obtain the probability of different categories, and classification loss is calculated through cross-entropy, which can be given as 
\begin{equation}
L\left( {r,p} \right) =  - \sum\limits_i {{r_i}\log {p_i}}
\label{eq:CE}
\end{equation}
where $i$ denotes the number of categories, $r$ and $p$ denotes the the ground truth and predicted probability, respectively. The classification loss measures the discrepancy between the true label and the prediction. With the training strategy based on backpropagation (BP) and optimization algorithms like stochastic gradient descent (SGD), the prediction of the model could approach the true label gradually.

\section{Methodology}
In this section, we introduce the structure and the kernel function of TFconv layer, present its interpreting method, and propose the TFN with the entire fault diagnosis procedure.

\subsection{Structure of TFconv Layer}
The TFT methods are physically interpretable, but they cannot extract adaptive time-frequency information based on the characteristics of the diagnostic dataset. On the other hand, CNNs can automatically extract high-dimensional features from the original samples and make accurate classifications efficiently, but the logic of CNN’s decision-making is not clear enough. To utilize the advantages of both two strategies, we embed the TFT method into the traditional convolutional layer since they can be both regarded as inner products, and this leads to a novel layer named the TFconv layer. 

\begin{figure}[htbp]
        \centering
        \includegraphics[width=8.5 cm]{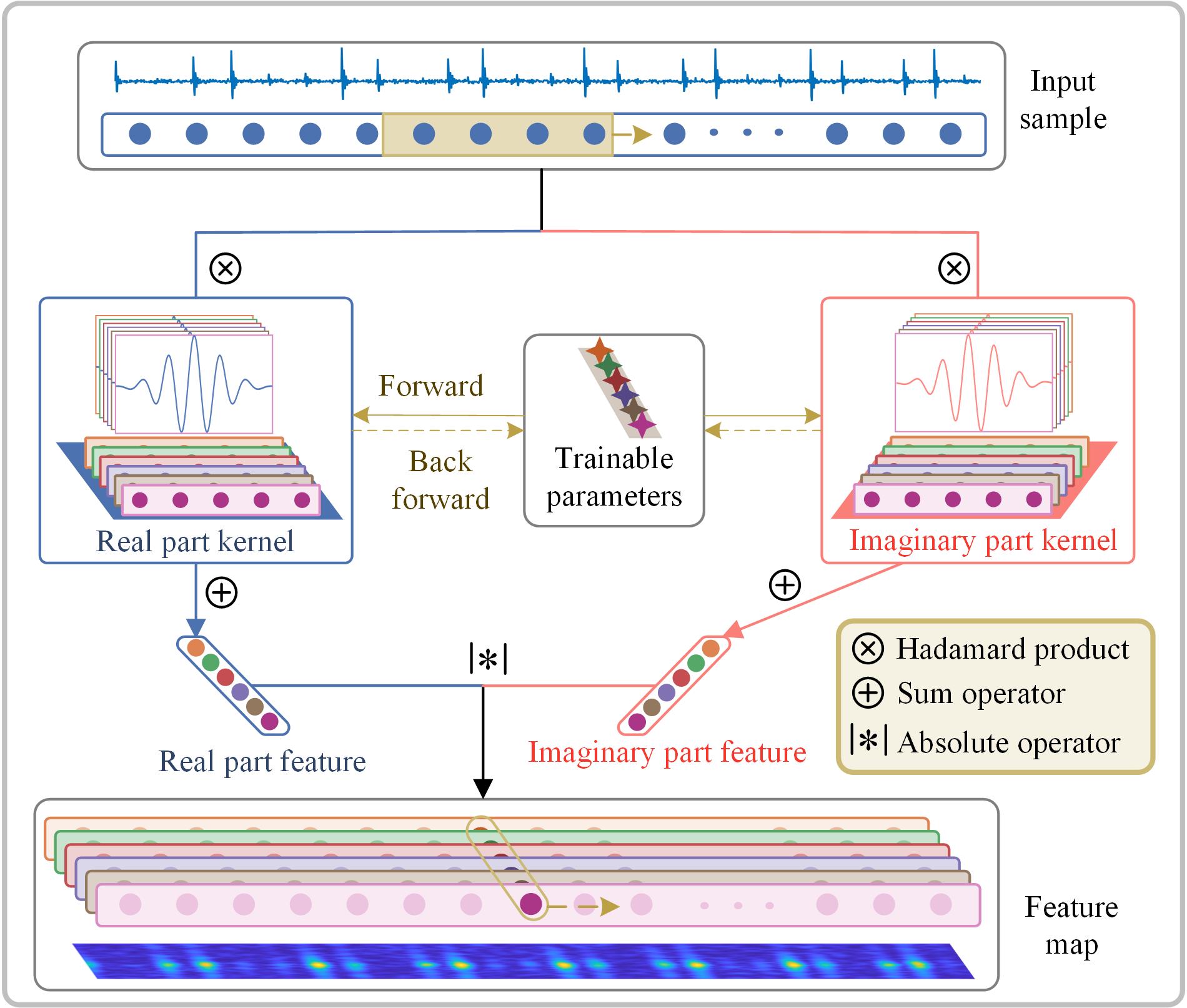}
        \caption{The process of TFconv layer.}
        \label{fig:TFconv process}
\end{figure}

 We designed the TFconv layer to simulate the physically interpretable TFT method, and the key point is to use a complex value convolutional kernel instead of the real value used in the current literature \cite{liWaveletKernelNetInterpretableDeep2022,ravanelliInterpretableConvolutionalFilters2019a,gangulyWaveletKernelBased2020}. Considering that most CNN models use real value variables and in order to get good compatibility, the structure of the TFconv layer is designed as Fig. \ref{fig:TFconv process}.  The TFconv layer consists of a real part kernel and an imaginary part kernel, which convolve with the input samples along the length direction to get real part features and imaginary part features, respectively. After that, the modulus of real and imaginary features is calculated (as TFT does) to obtain the final feature map as the output of the TFconv layer, and each channel is processed independently in the whole process.  Moreover, different from the traditional convolutional layers whose weights are randomly initialized, the weights of the real and imaginary part kernels are initialized and controlled be the kernel function, and their relationship can be expressed as
\begin{equation}
\left\{\begin{array}{l}
\begin{aligned}
\quad \psi_{\theta} \in \mathbb{C}, &\quad  \psi_{\theta, \text { real }}, \psi_{\theta, \text { imag }} \in \mathbb{R} \\
\psi_{\theta, \text { real }} &= \text{real}(\psi_{\theta})\\
\psi_{\theta, \text { imag }} &= \text{imag}(\psi_{\theta})
\end{aligned}
\end{array}\right.
\label{eq:TFconv real and imaginary part}
\end{equation}
where $\psi_{\theta}$ denotes the kernel function, $\psi_{\theta, \text { imag }}$ and $\psi_{\theta, \text { imag }}$ are the real part kernel and the imaginary part kernel respectively, ${\rm real}(\cdot)$ and ${\rm imag}(\cdot)$ denote the operators to get the real part and the imaginary part of a complex value respectively.  The control parameter $\theta$, which adjusts the frequency property of kernel function, is treated as the trainable parameter and updated in the BP process. The kernel function of the TFconv layer is equivalent to the inner product window function in TFT, and the output of the $k$-th channel of the TFconv layer can be denoted as
\begin{equation}
\left\{\begin{array}{l}
\begin{aligned}
h_{k, \text { real }}&=\psi_{\theta, \text { real }}^{k} * x \\
h_{k, \text {imag}}&=\psi_{\theta, \text {imag}}^{k} * x \\
h_{k}&=\sqrt{h_{k, \text { real }}^{2}+h_{k, \text {imag}}^{2}}
\end{aligned}
\end{array}\right.
\label{eq:TFconv}
\end{equation}
where $k$ denotes the $k$th channel of the convolutional kernel, $x$ denotes the input of the TFconv layer, $\theta$ denotes the trainable control parameter of the kernel function, $h_{\rm real}$ and $h_{\rm imag}$ denote the real and imaginary feature map respectively, and $h$ denotes the final feature map.

Compared to traditional convolutional layers, the proposed TFconv layer has three novel aspects as follows:
\begin{enumerate}
        \item Real-Imaginary Mechanism: TFconv layer has two convolutional processes of the real part kernel and imaginary part kernel, and their outputs are merged through modulo operation.
        \item Kernel Function: The weight of the convolution kernel of TFconv layer is determined by the specific kernel function, not randomly initialized.
        \item Trainable Parameters: The trainable parameters of TFconv layer are the control parameters $\theta$ of the kernel function (e.g., frequency factor $f$ in STFT kernel), instead of the convolutional weights.
\end{enumerate} 

Since the trainable parameters of the TFconv layer are different from those of the traditional convolutional layer, the BP process of the TFconv layer is different accordingly.  In the BP process, the TFconv layer calculates the gradient of trainable parameters and updates them during each training step, which can be expressed as
\begin{equation}
\left\{ \begin{array}{l}
{\delta _{{\theta}}} = \frac{{\partial L}}{{\partial {h}}}\left( {\frac{{\partial {h}}}{{\partial \psi _{p,{\rm real}}}}\frac{{\partial \psi _{p,{\rm real}}}}{{\partial {\theta}}} + \frac{{\partial {h}}}{{\partial \psi _{p,{\rm imag}}}}\frac{{\partial \psi _{p,{\rm imag}}}}{{\partial {\theta}}}} \right)\\
{\theta} \leftarrow \text{optimizer}\left({\theta}, {\delta _{{\theta }}}, \eta \right)
\end{array} \right.
\label{eq:TFconvBP}
\end{equation}
where $\theta$ denotes the trainable parameters of the kernel function, $\delta_{{\theta}}$ denotes the gradient of $\theta$, $\partial$ denotes the partial derivative operator, $L$ denotes the classification loss, $h$ denotes the output of the TFconv layer, $\psi _{p,{\rm real}}$ and $\psi _{p,{\rm imag}}$ denote the weights of the real and imaginary parts. After the gradient is obtained by the chain rule, the trainable parameters could be updated by optimization algorithms like SGD, Adam or RMSprop. 

\subsection{Kernel Function of TFconv Layer}

The kernel function of the TFconv layer derives from the inner product window function of TFT through necessary discretization and modification, which is the key step to embed the TFT method into the TFconv layer.  With the TFT method embedded, physical constraints of TFT are also introduced in the TFconv layer, that is, there is a certain limit on the trainable parameters of the kernel function. Taking STFT as an example, due to the \textit{nyquist} sampling theorem, the meaningful normalized frequency is $[0,0.5]$, so the frequency factor $f$ of the corresponding kernel function should be limited to $[0,0.5]$ as well.

According to the above analysis, three typical TFT methods (i.e., STFT, CT, and Morlet WT) are considered to formulate TFconv kernel functions. The inner product window functions of TFT, the corresponding kernel functions of the TFconv layer and trainable parameters with their limits are shown in Table \ref{tab:kernel Function}. Different from the short-time trigonometric function (STTF) and Chirplet function, the Morlet wavelet can stretch and contract in the time domain by scale factor $s$. To avoid time-domain truncation, the kernel length of the Morlet wavelet kernel function is designed to be much longer than that of the STTF and Chirplet kernel.

\begin{table}[htbp]
\caption{The Inner Product Functions of Time-Frequency Transform, The corresponding Kernel Functions of the TFconv layer and Trainable Parameter with Their Limits. ($N_c$ denotes the channel number of TFconv layer.)\label{tab:kernel Function}}
\centering
\resizebox{\textwidth}{!}{
\begin{tabular}{cccc}
\toprule[1pt]
Kernel Function & \makecell[c]{Inner Product Function of\\ Time-Frequency Transform} & \makecell[c]{Kernel Function of\\ TFconv layer} & \makecell[c]{Trainable Parameter\\$\theta$ with its Limit}\\
\midrule[0.5pt]
\vspace{3mm}
 STTF &
 \makecell[c]{$\psi_{\omega,\sigma}(t) = \frac{1}{\sqrt{2\pi}\sigma} e^{-\frac12 \left( \frac {t} \sigma \right)^2} \cdot e^{-j\omega t}$}&
  \makecell[c]{$\psi_{f}[n]=e^{-\frac{1}{2}\left(\frac{n}{\sigma N_c}\right)^{2}} e^{j 2 \pi f n}$,\\
  $\sigma=0.52$,\\
  $n=[-(N_c-1),\ldots,(N_c-1)]$}
    &$f_0 \in [0,0.5]$\\

 \vspace{3mm}
  Chirplet &
  \makecell[c]{$\psi_{\omega,\alpha,\sigma}(t) =\frac{1}{\sqrt{2\pi}\sigma} e^{-\frac12 \left( \frac {t} \sigma \right)^2} e^{-j\left[{\frac{\alpha}{2}t^2 + \omega t}\right]}$}&
  \makecell[c]{$\psi_{f, \alpha}[n]=e^{-\frac{1}{2}\left(\frac{n}{\sigma N_c}\right)^{2}} e^{-j 2 \pi\left[\frac{\alpha}{2} n^{2}+f n\right]} ,$\\
   $\sigma=0.52$,\\
  $n=[-(N_c-1),\ldots,(N_c-1)]$}&
   \makecell[c]{$f \in [0,0.5]$,\\$|\alpha|< 0.1/N_c $}\\
   \vspace{3mm}

   Morlet Wavelet &
    \makecell[c]{$\psi_{s}(t)=\frac{1}{\sqrt{s}} \Psi\left(\frac{t}{s}\right)$,\\
    $\Psi(t)=A e^{-\beta \frac{t^{2}}{2}} e^{j \omega t}$}&
    \makecell[c]{$\psi_{s}[n]=\frac{1}{\sqrt{s}} \Psi\left(\frac{n}{s}\right),$\\
    $\Psi(n)=e^{-\frac{1}{2}\left(\frac{n}{\sigma N_c}\right)^{2}} e^{j 2 \pi f_0 n}, $\\
    $\sigma=0.6,f_0=0.2$\\
    $n=[-10(N_c-1),\ldots,10(N_c-1)]$}&
    $s \in[0.4,10] $ \\
\bottomrule[1pt]
\end{tabular}}
\end{table}

\begin{figure}[htbp]
        \centering
        \includegraphics[width=12 cm]{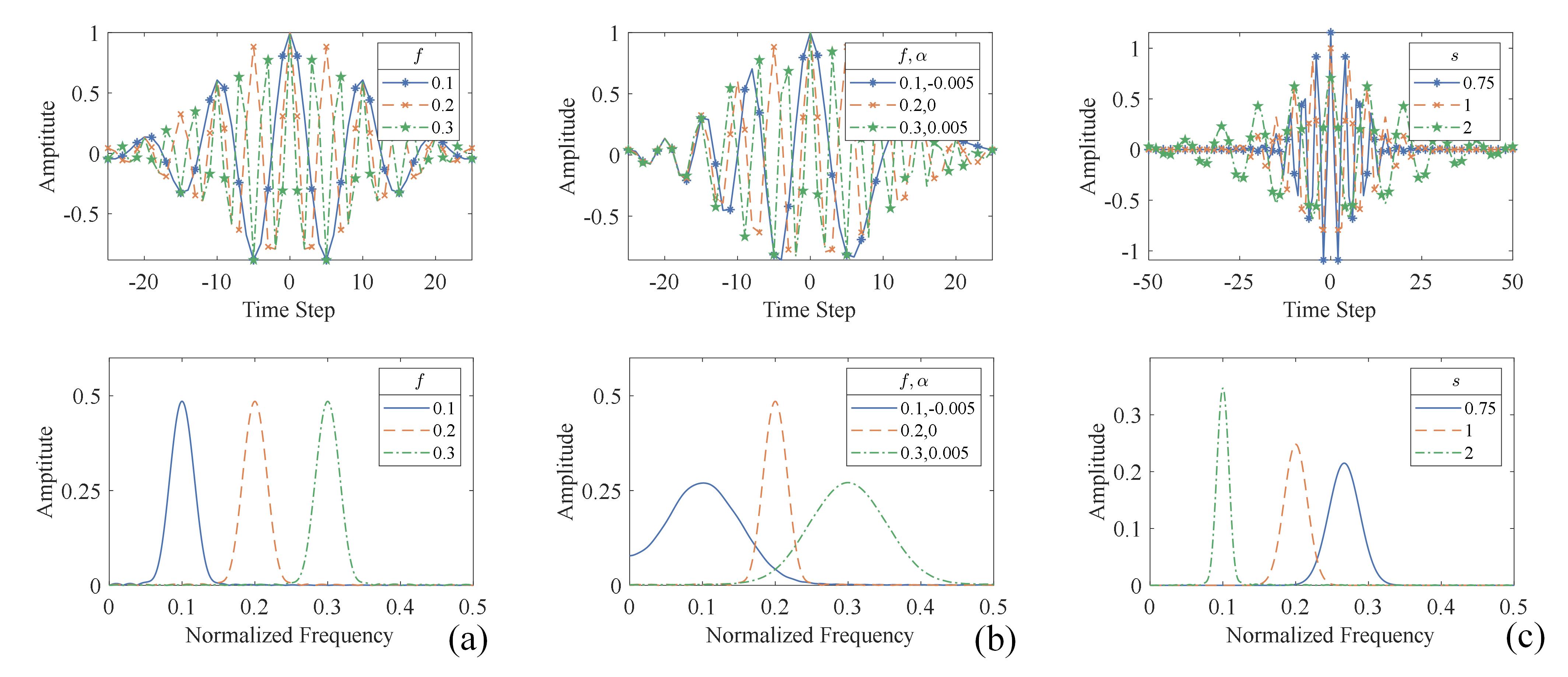}
        \caption{The time-domain and frequency-domain diagrams of  three  kernel functions of TFconv layer. (a) STTF. (b) Chirplet. (c) Morlet Wavelet.}
        \label{fig:kernel FR}
\end{figure}

To illustrate the properties of the above three kernel functions, the time-domain and frequency-domain diagrams of them are shown in Fig. \ref{fig:kernel FR}, and all the kernel functions can be regarded as bandpass filters. STTF kernel function has a fixed frequency bandwidth, and the center frequency is adjusted by the frequency factor $f$. The Chirplet kernel function introduces a linear frequency modulation factor $\alpha$ to the STTF kernel function and can dynamically change its filtering bandwidth. Morlet  Wavelet kernel function can scale their mother wavelet by scaling factor $s$ to adjust their frequency properties, and with the increase of scaling factor, it has a higher center frequency and wider bandwidth.

\subsection{Interpretability of TFconv Layer}
With the TFT method embedded, the TFconv layer can not only extract time-frequency information from input samples, but also become explainable through the interpreting method proposed in this subsection. The interpreting method is to perform the frequency response analysis on the trained TFconv layer to obtain its amplitude-frequency response (FR), which indicates the attention that CNN paid to different frequencies. The frequencies with higher attention could easily pass through the TFconv layer and are deeply involved in the prediction-making process and responsible for the results of CNN.  

The frequency response analysis of the convolutional layer derives from the filter theory in the signal processing field. The convolutional process in the convolutional layer is exactly equivalent to the filtering process of the finite impulse response (FIR) filter \cite{andrearczykUsingFilterBanks2016}. The convolution kernel can be regarded as the FIR filter, and the input sample is the signal to be filtered. In the frequency response analysis, the convolution kernel is processed by fast Fourier transform (FFT) to obtain the FR of the convolutional layer \cite{oppenheim1997signals}. Considering multiple channels contained in the convolutional layer, we calculate the channel-wise amplitude-frequency response (C-FR) first, and average them to obtain the overall amplitude-frequency response (O-FR). The calculation process could be denoted as
\begin{equation}
\begin{split}
{H_i}(f) &= \left| {\rm FFT}(w_i) \right| = \left| \sum\limits_{n=0}^{N}{w_i[n]e^{-j\frac{2\pi fn}{N}}} \right|\\
H(f) &= \frac{1}{n_c}\sum\limits_i^{n_c} {{H_i}(f)} 
\end{split}
\label{eq:FR}
\end{equation}
where, $w_i$ denotes the convolution kernel of the $i$th channel, $H_i(f)$ and $H(f)$ denote the C-FR of the $i$th channel and the O-FR, respectively. Although the above frequency response analysis is based on the traditional convolutional layer, it is also applicable to the TFconv layer because the real part kernel and imaginary part kernel in the TFconv layer are equivalent to a complex value kernel.

\begin{figure}[htbp]
        \centering
        \includegraphics[width=12 cm]{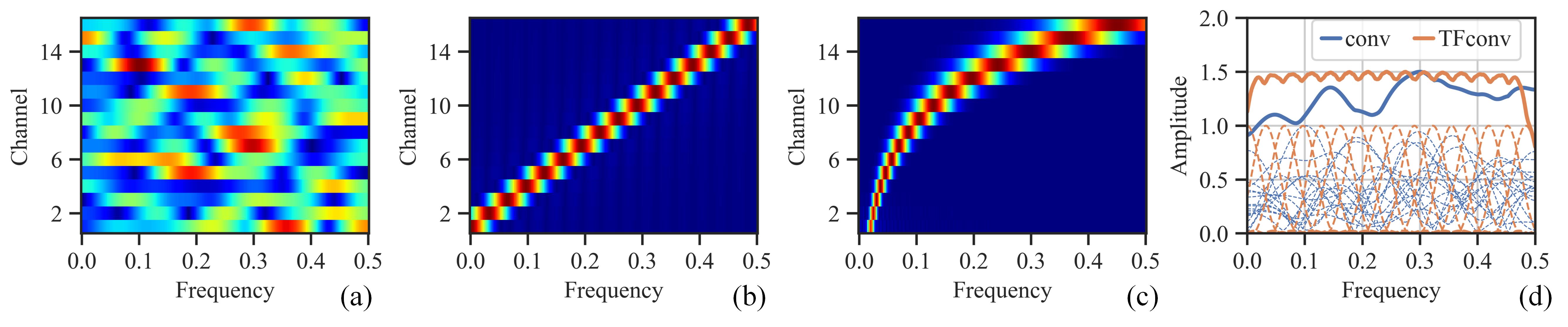}
        \caption{The comparison between  C-FR and O-FR of initialized traditional convolutional layer and that of initialized TFconv layers. (a) C-FR of traditional convolutional layer. (b) C-FR of TFconv layer with STTF kernel.  (c) C-FR of TFconv layer with Morlet wavelet kernel.  (d) O-FR of traditional convolutional layer and TFconv layer with STFT kernel (solid line represents O-FR while dashed line represents C-FR).}
        \label{fig:Comparison FR}
\end{figure}

Both the traditional convolutional layer and the TFconv layer can be conducted by frequency response analysis, yet there still exist some differences in their results.
The C-FR of the traditional convolutional layer, shown in Fig. \ref{fig:Comparison FR} (a),  presents a random distribution, which is difficult to identify the focusing frequency area (i.e., the passband frequency of the FIR filter).  While the TFconv layer, shown in Fig. \ref{fig:Comparison FR} (b) and (c), is controlled by the kernel function and has a clear frequency preference. Without considering the modulo operation, the TFconv layers are equivalent to a series of bandpass FIR filters, and different channel focuses on a different frequency.  The O-FR of the traditional convolutional layer and TFconv layer are shown in Fig. \ref{fig:Comparison FR} (d). Compared with the traditional convolutional layer, the focusing frequency area of the TFconv layer can be identified more easily and used for interpreting the frequency foundation of CNN models.

 Moreover, it is necessary to further elaborate on the interpreting logic of the TFconv layer. TFconv layer can be regarded as a series of trainable bandpass FIR filters, without considering the modulo operation. When the TFconv layer is combined with a backbone CNN as a preprocessing layer, the TFconv layer becomes the data portal of the backbone CNN. This means that only the frequency with amplitude in the amplitude-frequency response (O-FR) can pass the TFconv layer and be used by the subsequent backbone CNN for fault diagnosis tasks. Therefore, we claim that our TFN models have interpretability because the overall amplitude-frequency response (O-FR) of the TFconv layer could be used to explain the attention that the CNN model paid to different frequencies. To validate this, we propose a hypothesis that, the information frequency bands (which carry most information of the dataset) are crucial for fault diagnosis, and TFN models will pay more attention (i.e., amplitude peaks in O-FR) to these information frequency bands during the training process to achieve better diagnostic performance. We will verify the above hypothesis in the following experiment section, and if the O-FRs of trained TFN models have a good correspondence with the information frequency bands of the dataset frequency spectrum, the correctness of the interpreting logic of proposed TFN models would be effectively confirmed.  

\subsection{Fault Diagnosis Using TFN}

\begin{figure*}[!t]
        \centering
        \includegraphics[width=\textwidth]{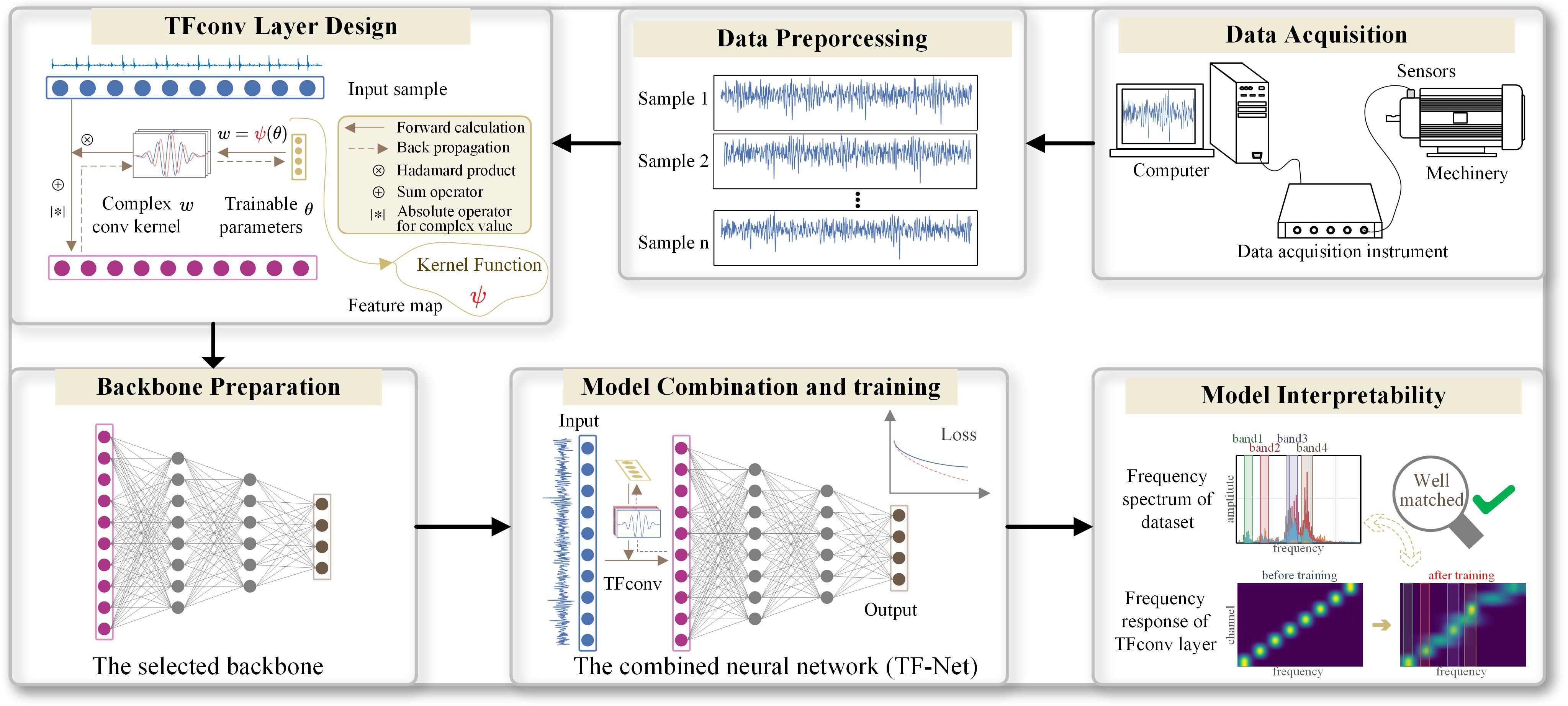}
        \caption{The entire process of applying TFN to intelligent mechanical fault diagnosis}
        \label{fig:framwork}
\end{figure*}

The interpretable TFconv layer is used as a preprocessing layer to combine with a backbone  CNN, and the resulting novel network is named the Time-Frequency Network (TFN). With TFN, we can extract fault-related time-frequency information from raw vibration signals and diagnose the fault states of mechanical equipment efficiently. The entire process of applying TFN to intelligent mechanical fault diagnosis is summarized in Fig. \ref{fig:framwork}.

Firstly, the vibration signals are collected by the accelerometers installed on the mechanical equipment. Secondly, the vibration signal is truncated with a sliding window to generate a series of samples as the input of TFN. Thirdly, the kernel function is formulated from TFT methods and embedded into the TFconv layer. Then, an existing CNN model is selected as the backbone, and the TFconv layer is combined as the preprocessing layer with the backbone to obtain TFN. After that, the obtained TFN is trained and verified by training samples and testing samples in fault diagnosis tasks. Finally, the interpreting method is conducted to reveal the focusing frequency area of TFN.

In order to verify the effectiveness of the TFconv layer, a relatively shallow CNN is selected as the backbone, and the architecture of the obtained TFN is shown in Table \ref{tab:TFN Structure}. $n_c$, $N$ and $n_p$ are the channel number of the preprocessing layer, the length of the TFconv layer, and the number of categories for classification, respectively. $n_c$ and $N$ are determined in the design process of the TFconv layer, and $n_p$ is determined by the specific dataset.

\begin{table}[htbp]
\caption{The Architecture of The Analyzed TFN\label{tab:TFN Structure}}
\centering
\begin{tabular}{ccccc}
\toprule[1pt]
 Part & No.~Unit & Basic Unit & Output Size\\
\midrule[0.5pt]
\vspace{1mm}
 TFconv& - &  input & 1*1024 \\
 
  & 1 & TFconv($n_{\rm c}$@N*1) &$n_{\rm c}$*1024\\
 \midrule[0.5pt]
 \vspace{1mm}
 Backbone & 2 &  Conv(16@15*1)-BN-ReLU & 16*1010 \\
 \vspace{1mm}
   & 3 & Conv(32@3*1)-BN-ReLU-MaxPool(2) & 32*504 \\
   \vspace{1mm}
    & 4 & Conv(64@3*1)-BN-ReLU & 64*502 \\
    \vspace{1mm}
     & 5 & Conv(128@3*1)-BN-ReLU-AdaptivePool(4) & 128*4\\
     \vspace{1mm}
      & 6 & Flatten & 512 \\
        & 7 & Dense(256)-ReLU-Dense(64)-ReLU-Dense($n_{\rm p}$) & $n_{\rm p}$ \\
\bottomrule[1pt]
\end{tabular}
\end{table}

\section{Experiment}

In this section, three experimental datasets are used to verify the diagnostic performance and  interpretability of TFN.  In the diagnostic performance part, considering the importance of the channel numbers, we compare TFN models with other models mentioned in Table \ref{tab:comparison with our work} under different channel numbers.  In the interpretability part, the consistency between the frequency spectrum of the dataset and the O-FRs of trained TFconv layers with different kernels are analyzed. 

\subsection{CWRU Public Bearing Dataset}

\begin{figure}[htbp]
        \centering
        \includegraphics[width=6 cm]{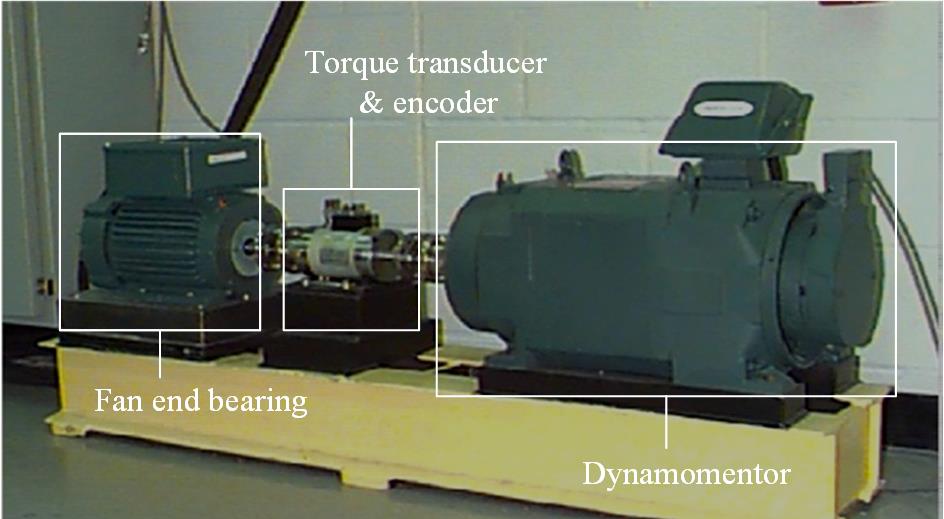}
        \caption{CWRU bearing experimental system.}
        \label{fig:CWRU rig}
\end{figure}

In order to better spread our work, we choose this open-source dataset as a benchmark to test the diagnostic performance of TFNs. The CWRU bearing dataset \cite{smithRollingElementBearing2015}, as shown in Fig. \ref{fig:CWRU rig}, is one of the most popular open-source datasets for mechanical fault diagnosis. The accelerometer is installed on the motor casing of the drive end, and the vibration signals are collected under four loading conditions (load 0-3 HP) at two sampling frequencies of 12 kHz and 48 kHz. 

In addition to normal condition (N), three different bearing fault types are contained in this dataset: inner raceway fault (I), rolling element fault (B), and outer raceway fault (O). For each fault type, different fault sizes were considered respectively, i.e., 0.007, 0.014 and 0.021 inches. Therefore, this dataset has nine fault states and a normal state, ten categories in total. The fault diagnosis of CWRU rolling bearing can be regarded as a 10-class classification.

The diagnostic difficulty of the CWRU dataset is relatively low, and the diagnostic accuracy of 12 kHz signals of most models is close to 100\%, which is hard to distinguish the diagnostic performance of different models. Therefore, the 48 kHz vibration signal is chosen for the following diagnosis task for its higher diagnostic difficulty. The vibration signal is truncated into samples at the length of 1024 and each state has 450 samples, for a total of 4500 samples. 60\% of samples in each category are used for training, and the remaining samples are used for testing. The loss function of classification is cross-entropy, the training optimizer is Adam, the momentum is set to 0.9, the initial learning rate is 0.001, the decay ratio of the learning rate is 0.99 per epoch, and the training epoch is 50. Each model is repeated 10 times to eliminate randomness and verify the stability.

To test the diagnostic performance of the TFconv layer, this experiment contains three types of models. 1) Backbone models: the backbone CNN, the backbone CNN with traditional random kernel convolutional layer as a preprocessing layer (denoted as Random-CNN). 2) Contrast models: SincNet\cite{ravanelliInterpretableConvolutionalFilters2019a}, WKN with Morlet kernel and Laplace kernel(denoted as WKN-Morlet and WKN-Laplace)\cite{liWaveletKernelNetInterpretableDeep2022}, W-CNN\cite{gangulyWaveletKernelBased2020}. 3) TFN models: TFNs with STTF, Chirplet, and Morlet Wavelet as kernel functions (denoted as TFN-STTF, TFN-Chirplet, and TFN-Morlet, respectively). Except for the backbone CNN, each type of other model contains 4 different channel numbers of preprocessing layer: 16, 32, 64 and 128.

\begin{figure}[H]
        \centering
        \includegraphics[width=15 cm]{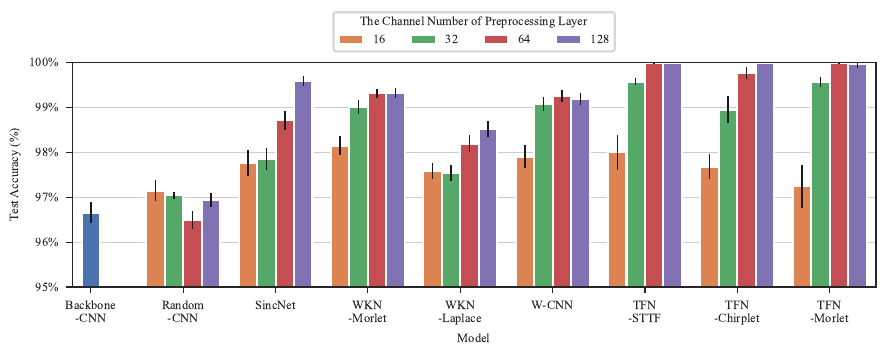}
        \caption{Test accuracy on the CWRU bearing dataset with different loading conditions.}
        \label{fig:CWRU acc}
\end{figure}

The fault diagnosis results of the CWRU bearing dataset are shown in Fig. \ref{fig:CWRU acc}, and we can draw the following conclusions.

\begin{enumerate}
        \item The diagnostic accuracies of Backbone-CNN and Random-CNN are the worst. SincNet, WKN-Morlet, WKN-Laplace, and W-CNN have made great improvements based on Backbone CNN. Furthermore, TFN-STTF, TFN-Chirplet and TFN-Morlet reach the best diagnostic performance overall, and their diagnostic accuracy under 64 or 128 channel numbers is close to 100\%, demonstrating the effectiveness of TFN in fault diagnosis. The TFconv layer of TFN can transform the raw vibration signal into fault-related time-frequency features, leading to the distinguished diagnostic performance of TFN.

        \item The more channels generally led to the higher diagnostic accuracy and this phenomenon is more obvious in TFN models. The diagnostic accuracy of TFN models and other models are similar when the channel number is 16, but with the increase of the channel number, the diagnostic accuracies of TFN models are significantly higher than that of other methods. TFconv layer with more channels could extract more time-frequency information and lead to higher diagnostic accuracy. However, the diagnostic accuracy of TFN with 64 channels is similar to that with 128 channels. This indicates that when the TFconv layer has enough channels to extract time-frequency information, increasing the number of channels does not bring a corresponding increase in diagnostic accuracy.

        \item As for the kernel functions of TFN, all three kernel functions have similar diagnostic performance, and the STTF kernel is slightly better than the Chirplet kernel and Morlet kernel in general.
\end{enumerate}

After introducing the diagnostic performance of TFN, here comes the interpretability analysis part. To have a suitable dataset frequency spectrum, the vibration signals at 12 kHz sampling frequency under 3HP load condition in the CWRU dataset are used as the input samples, and TFNs with different kernel functions are trained by them. In order to obtain a clear interpretable observation, the channel numbers of preprocessing layer are set to 8 and other experimental settings are the same as those in the previous diagnostic experiments.

\begin{figure}[htbp]
        \centering
        \includegraphics[width=15.6 cm]{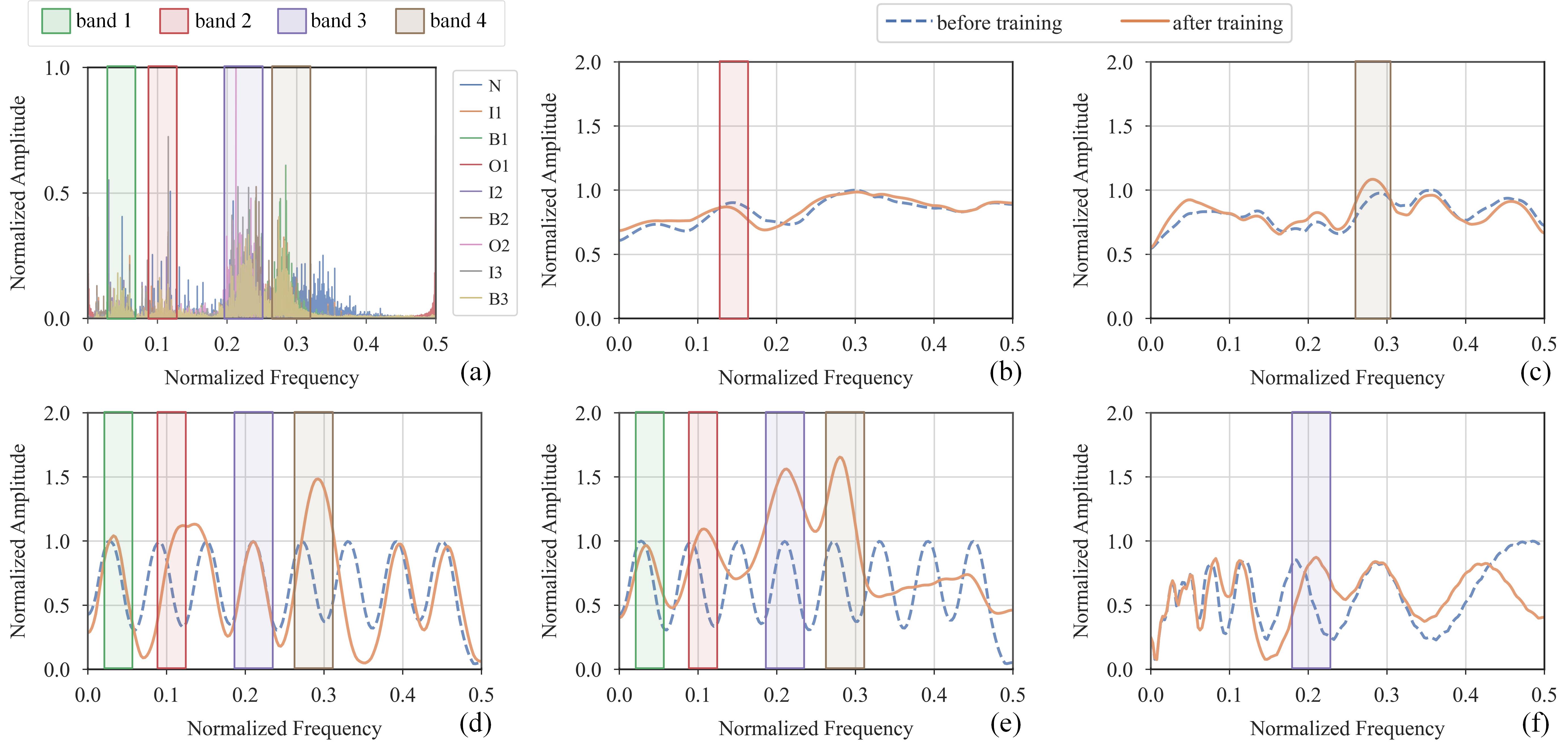}
        \caption{Frequency spectrum of the CWRU bearing dataset and O-FRs of different models. 
        (a) Frequency spectrum. (b) O-FR of the first convolutional layer of Backbone-CNN.
        (c) O-FR of the first convolutional layer of Random-CNN. (d) O-FR of the TFconv layer of TFN-STTF. 
        (e) O-FR of the TFconv layer of TFN-Chirplet. (f) O-FR of the TFconv layer of TFN-Morlet. }
        \label{fig:interpretability}
\end{figure}

According to frequency response analysis (Eq.\ref{eq:FR}), the O-FR of the preprocessing layer of different well-trained models could be obtained. The frequency spectrum of CWRU dataset and the O-FR results are shown in Fig. \ref{fig:interpretability}. As shown in the frequency spectrum of the CWRU dataset (Fig. \ref{fig:interpretability}(a)), the frequency amplitude mainly exists in four information frequency bands, i.e., bands 1-2-3-4, which carry most information of the dataset and have been marked out by different colors. These information bands are responsible for fault diagnosis and a good CNN model should focus on them to achieve a satisfying diagnostic accuracy. From the O-FR results shown in Fig. \ref{fig:interpretability}(b)-(f), the following conclusions could be obtained.
\begin{enumerate}
        \item The O-FRs of trained Backbone-CNN and Random-CNN have an amplitude increase only at band 2 or band 4, indicating that the information is insufficiently extracted for fault diagnosis.
        \item The O-FR of trained TFN-STTF has amplitude peaks at all the information frequency bands. It means that TFN-STTF pays correct attention to these information bands, which is consistent with its outstanding performance in fault diagnosis. However, since the parameter $f$ of the STTF kernel function can only change the centering frequency, not the bandwidth, there still exist some amplitude peaks outside the information bands (i.e., two amplitude peaks in high-frequency area).
        \item The O-FR of trained TFN-Chirplet has amplitude peaks within all the information frequency bands as well.  However, different from the STTF kernel function, the Chirplet kernel function has an additional linear frequency modulation factor $\alpha$ to change its filtering bandwidth, and it would increase its bandwidth to search a wider frequency band when this channel can not get any useful information, so there exists no amplitude peak outside the information bands in the OF-R of TFN-Chirplet.  The O-FR of TFN-Chirplet is completely consistent with the frequency spectrum of the CWRU dataset, and TFN-Chirplet achieves the best physical interpretability than other models.
        \item The O-FR of TFN-Morlet changes little after training, where only one amplitude peak (corresponding to band 3) can be barely identified. Although the adaptive frequency bandwidth of the wavelet can help to extract fault-related information, it also blurs the focusing frequency bands and thus prevents them from being identified. This leads to the inferior interpretability of TFN-Morlet compared with TFN-STTF and TFN-Chirplet.
\end{enumerate} 

By analyzing the O-FR of the trained TFconv layer, we confirm the excellent performance of the proposed TFNs in terms of interpretability. TFN-Chirplet has the best interpretability, followed by TFN-STTF. Although TFN-Morlet has an excellent performance in fault diagnosis, the interpretability result of TFN-Morlet is not as good as that of TFN-Chirplet and TFN-STTF.

\subsection{Planetary Gearbox Dataset}

\begin{figure}[htbp]
        \centering
        \includegraphics[width=9 cm]{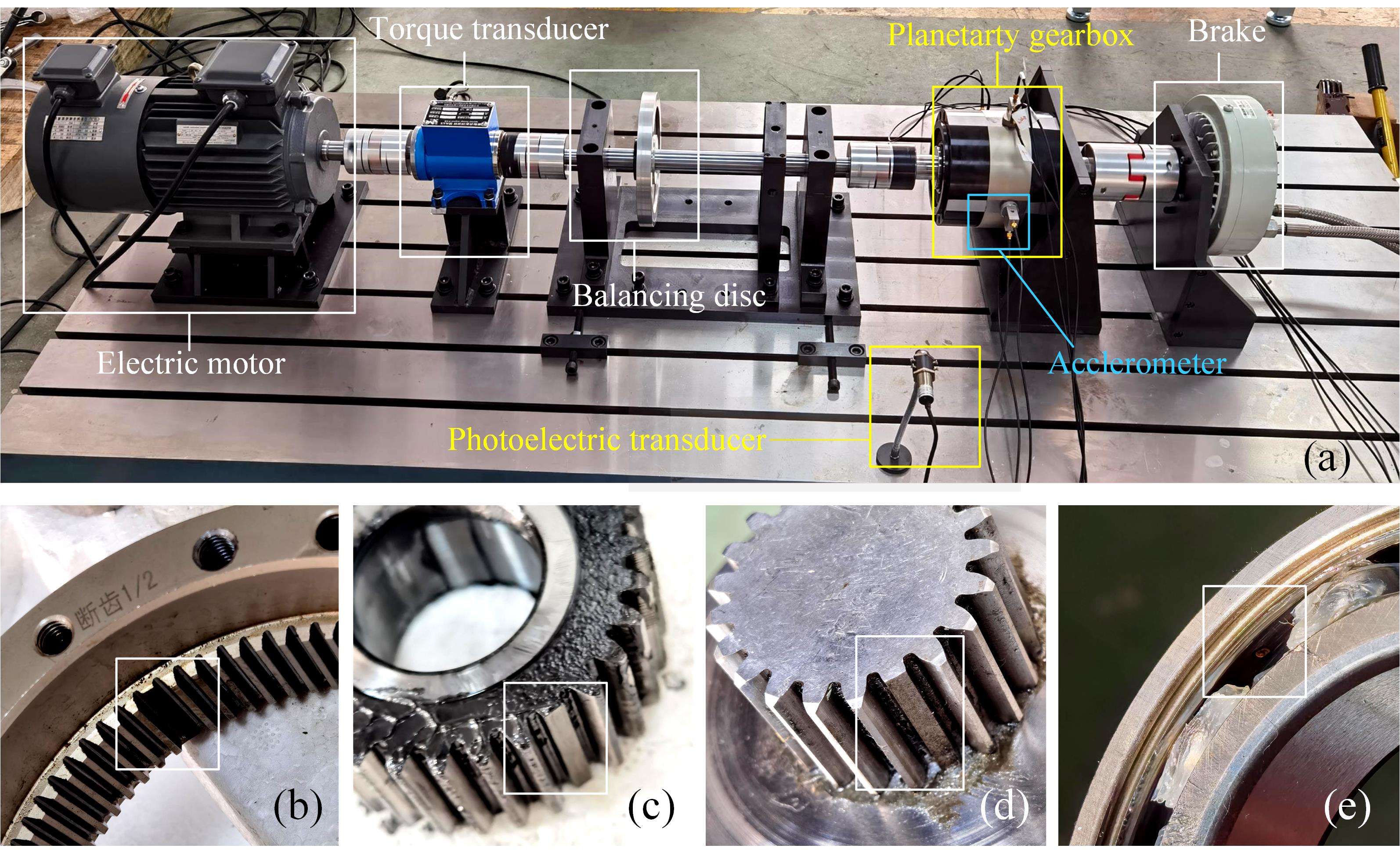}
        \caption{The planetary gearbox system and the defective components. (a) The planetary gearbox system. (b)Tooth fracture of the ring gear. (c) Tooth fracture of the sun wheel. (d) Tooth fracture of the planetary gear. (e) Outer race pitting of the rolling bearing.}
        \label{fig:houde rig}
\end{figure}

As shown in Fig. \ref{fig:houde rig} (a), this planetary gearbox system includes an electric motor, a transmission shaft, a torque transducer, a planetary gearbox, a magnetic powder brake and a series of sensors. The vibration signals are collected by the accelerometer located at the shell of the planetary gearbox and transmitted to the signal acquisition card at the sampling frequency of 10.2 kHz.

 Four types of component failures are considered in this dataset, including tooth fractures of three different gears and the outer race pitting of the rolling bearing as shown in Fig. \ref{fig:houde rig} (b)-(e). Based on these faults, the experiments are conducted under five health conditions listed in Table \ref{tab:houde condition}, specifically normal state (N), single-point fault (S), double-point fault (D), three-point fault (T) and compound fault (C). The fault diagnosis of the planetary gearbox can be regarded as a 5-class classification task.

\begin{table}[htbp]
\caption{The Working Condition of Gearbox System\label{tab:houde condition}}
\centering
\begin{tabular}{ccc}
\toprule[1pt]
Label & Failure Component & Training/Testing sample\\
\midrule[0.5pt]
N     & None                                      & 264/176                \\
S     & Ring   Gear                               & 264/176                \\
D     & Ring   Gear and Sun Wheel                 & 264/176                \\
T     & Ring   Gear, Sun Wheel and Planetary Gear & 264/176                \\
C     & Ring   Gear and Rolling Bearing           & 264/176 \\
\bottomrule[1pt]
\end{tabular}
\end{table}

In data preparation, the raw vibration signal is truncated without overlap through a sliding window to obtain input samples. Each category contains 440 samples, and the length of each sample is set to 1024. After that, 60\% of the samples are randomly divided as the training set, and the rest samples are used as the test set. In addition, gaussian white noise with signal to noise ratio (SNR) of 0 is added to the original signal to increase the difficulty of diagnosis. The rest of the experimental settings are consistent with that of the diagnostic experiment on the CWRU dataset.

\begin{figure}[htbp]
        \centering
        \includegraphics[width=15 cm]{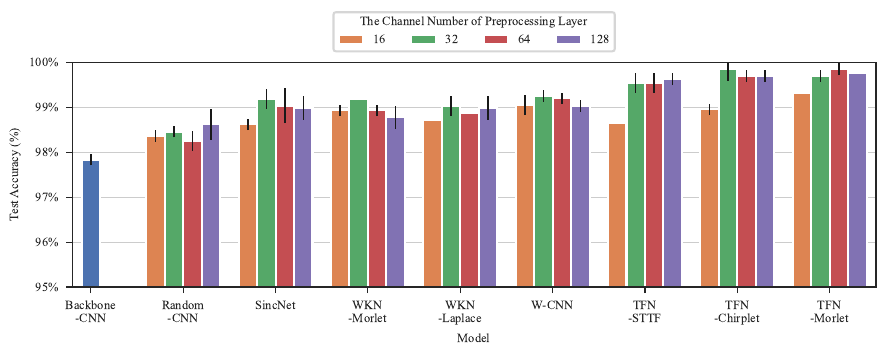}
        \vspace{-2 mm}
        \caption{Test accuracy on the planetary gearbox dataset.}
        \label{fig:houde acc}
\end{figure}

The diagnostic results are shown in Fig. \ref{fig:houde acc}. The diagnostic difficulty of this dataset is relatively low, and the diagnostic performance gap of different models is not as obvious as in the previous CWRU diagnostic experiment. The diagnostic accuracy of Backbone-CNN is 97.8\%, and Random-TFN performs better than the backbone CNN. SincNet, WKN-Morlet, WKN-Laplace, and W-CNN perform slightly better than the previous two models. TFN-STTF, TFN-Chirplet and TFN-Morlet achieve the best overall diagnostic performance, demonstrating the effectiveness of the proposed method. Furthermore, TFNs with 32 channels perform much better than that with 16 channels, but TFNs with more channels (i.e., 64, 128) do not have a corresponding increase in accuracy. It shows that 32 channels are enough for TFN to extract sufficient Time-Frequency information on this planetary gearbox dataset.

\begin{figure}[htbp]
        \centering
        \includegraphics[width=15.6 cm]{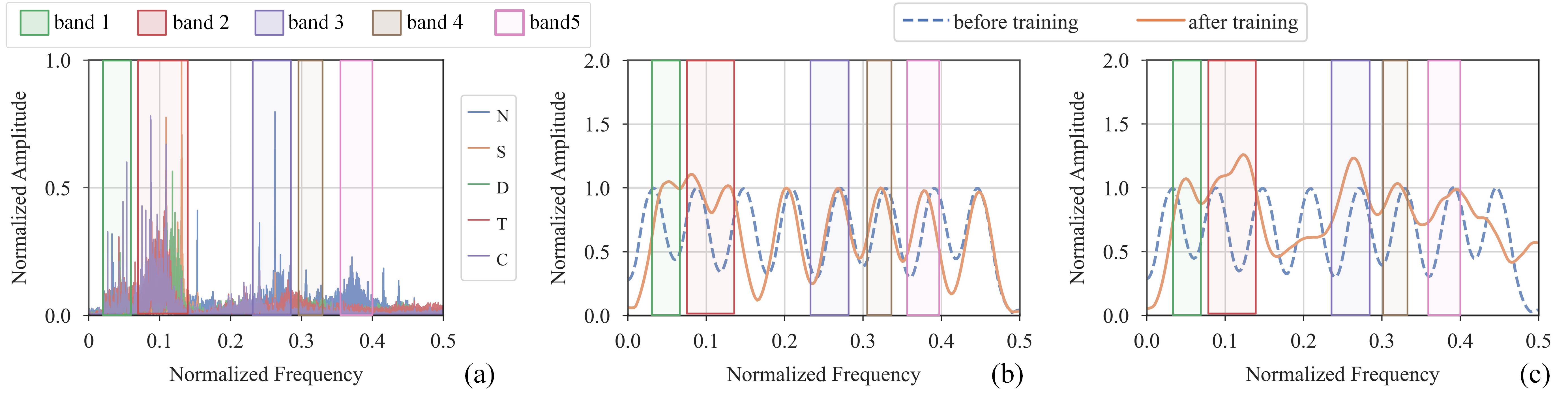}
        \caption{ Frequency Spectrum of the planetary gearbox dataset and O-FRs of different models. 
        (a) Frequency Spectrum. (b) O-FR of the TFconv layer of TFN-STTF. (c) O-FR of the TFconv layer of TFN-Chirplet. }
        \label{fig:interpretability-Houde}
\end{figure}

As for the interpretability analysis on the planetary gearbox dataset, considering the inferior interpretability performance of backbone models and TFN-Morlet, we only present the interpretability results of TFN-STTF and TFN-Chirplet for brevity. The channel numbers of the preprocessing layer are set to 8 and other experimental settings are the same as those in the previous diagnostic experiment on the planetary gearbox dataset. The frequency spectrum of the planetary gearbox dataset and the O-FRs of TFN-STTF and TFN-Morlet are shown in Fig. \ref{fig:interpretability-Houde}.

As marked in Fig. \ref{fig:interpretability-Houde}(a),  the frequency amplitude mainly exists in five information frequency bands that contain most information of the planetary gearbox dataset, i.e., bands 1-2-3-4-5. As shown in Fig. \ref{fig:interpretability-Houde}(b)-(c), both the O-FRs of trained TFN-STTF and trained TFN-Chirplet have amplitude peaks at these five information frequency bands, showing that TFN-STTF and TFN-Chirplet pay correct attention to the information frequency bands of planetary gearbox dataset. However, the Chirplet kernel function has an additional linear frequency modulation factor $\alpha$ for adjusting its filtering bandwidth compared to the STTF kernel function, so the O-FR of TFN-STTF has two amplitude peaks outside the information frequency band (one is nearly 0.1 in frequency and the other is nearly 0.45 in frequency), while  the O-FR of TFN-Chirplet has no unrelated amplitude peaks and is more consistent with the frequency spectrum of the planetary gearbox dataset. In conclusion, this phenomenon in the planetary gearbox dataset is the same as that of the CWRU bearing dataset, demonstrating the outstanding interpretability of TFN-Chirplet to reveal the focusing frequency area of CNN models.

\subsection{Aerospace Bearing Dataset}

\begin{figure}[H]
        \centering
        \includegraphics[width=12 cm]{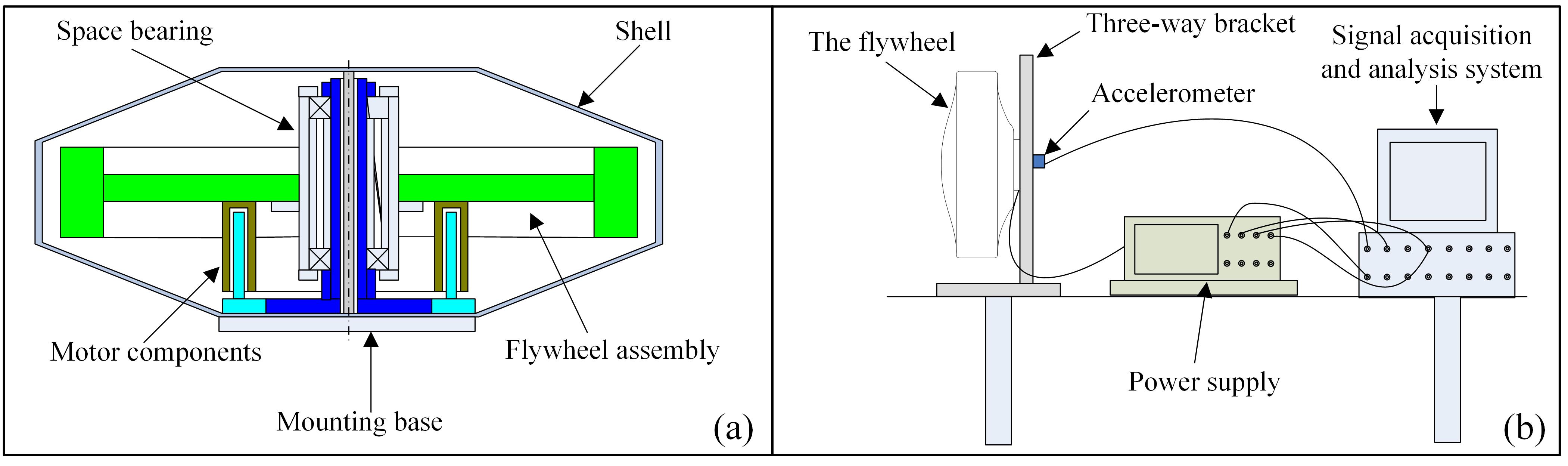}
        \vspace{-2 mm}
        \caption{The configuration of aerospace bearing experimental system. (a) Structure of the flywheel. (b) Schematic of the flywheel vibration acquisition process.}
        \label{fig:aerobearing rig}
\end{figure}

The former two datasets are collected in laboratory scenarios, while this aerospace bearing dataset is derived from the industrial scenario. As shown in Fig. \ref{fig:aerobearing rig}, the aerospace bearing is the core component of the flywheel test rig, which consists of motor components, an aerospace bearing, a flywheel assembly, a shell, and a mounting base as shown in Fig. \ref{fig:aerobearing rig}(a). The flywheel is driven by an electric motor and then puts the aerospace bearing to work. The data acquisition equipment is shown in Fig. \ref{fig:aerobearing rig}(b), including an accelerometer, power supply, a signal acquisition and analysis system. The flywheel is fixed on an upright bracket, where the acceleration sensor collects the three-way vibration signals of the aerospace bearing at the sampling frequency of 25.6 kHz.

The aerospace bearing dataset has five types of health conditions: normal state (N), leading surface scratching (S), cage fault (C), ball fault (B), and inner ring fault (I). For each condition, the radial vibration signal is truncated into samples, and each category contains 1000 samples, for a total of 5000 samples. Then 60\% of the samples in each category are used for training, and the remaining samples are used for testing. The fault diagnosis of the aerospace bearing dataset can be regarded as a 5-class classification task. Like the processing of the planetary gearbox dataset, the Gaussian noise of SNR = 0 is added to the original signal to increase the difficulty of diagnosis. The rest settings for this experiment are consistent with that of the previous planetary gearbox experiment.

\begin{figure}[H]
        \centering
        \includegraphics[width=15 cm]{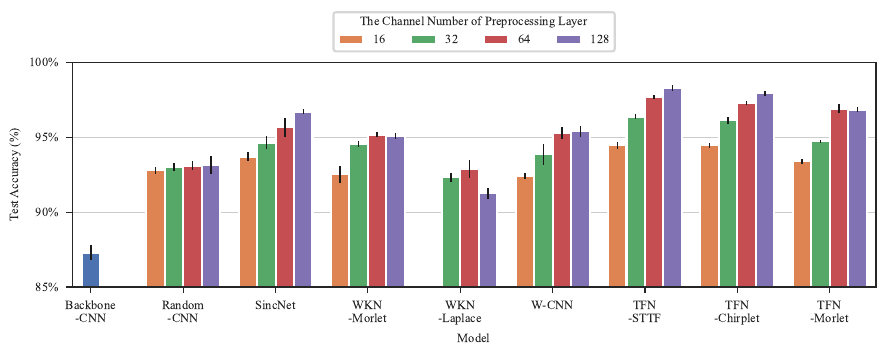}
        \vspace{-2 mm}
        \caption{Test accuracy on the aerospace bearing dataset.}
        \label{fig:aerobearing acc}
\end{figure}

The experimental results are shown in Fig. \ref{fig:aerobearing acc}. The performance of the Backbone CNN (87.3\%) is the worst and Random-TFN (around 93\%) performs much better. This may be caused by the increased model depth of the newly added convolutional layer, even it is randomly initialized and updated through BP process. The contrast type models (including SincNet, WKN-Morlet, WKN-Laplace, W-CNN) is close or slightly better than Random-CNN, and the diagnostic accuracy of SincNet with 128 channels is 96.7\%. As for TFNs, TFN-STTF, TFN-Chirplet, and TFN-Morlet have best performances on diagnostic accuracy, where STTF-TFN with 128 channels achieves the highest average accuracy of 98.3\%. TFconv layer with more channels could extract more time-frequency information and lead to a higher diagnostic accuracy, but at the expense of training time. In conclusion, proposed TFNs have a much better performance than both backbone models and contrast models, and the diagnostic accuracy of TFNs increase significantly with the increase of the number of channels, demonstrating the importance of the number of channels.

\begin{figure}[htbp]
        \centering
        \includegraphics[width=15.6 cm]{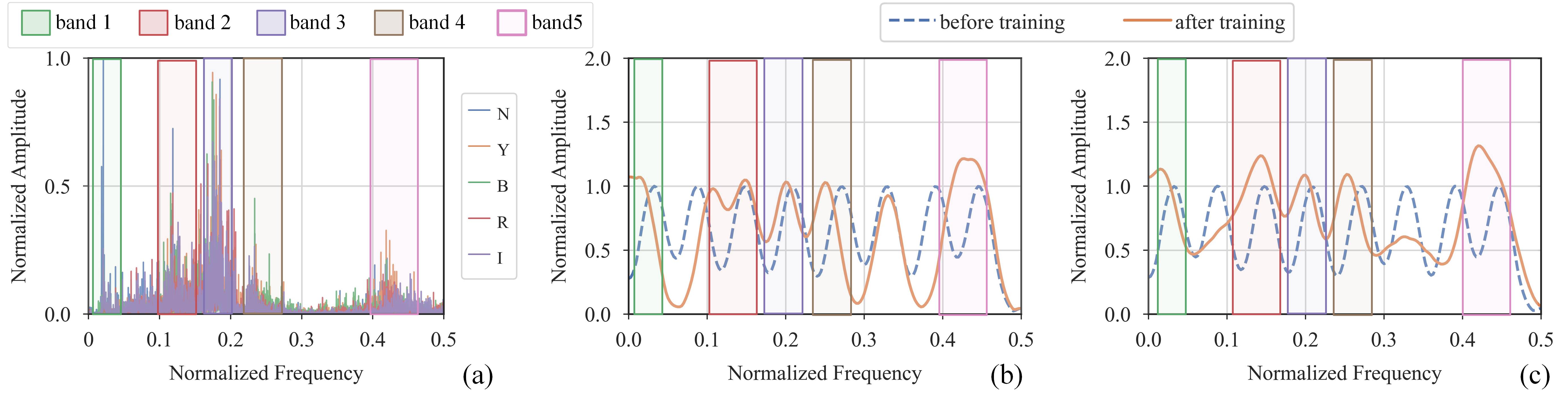}
        \caption{ Frequency Spectrum of the aerospace bearing dataset and O-FRs of different models. 
        (a) Frequency Spectrum. (b) O-FR of the TFconv layer of TFN-STTF. (c) O-FR of the TFconv layer of TFN-Chirplet. }
        \label{fig:interpretability-aerobearing}
\end{figure}

As for the interpretability analysis, the process on the aerospace bearing dataset is the same as the previous two datasets. Specifically, TFN-STTF and TFN-Morlet with 8 preprocessing layer channels are chosen to be trained on the aerospace bearing dataset, and the experiment settings are the same as the diagnostic experiment before. The frequency spectrum of the aerospace bearing dataset and the O-FRs of TFN-STTF and TFN-Morlet are shown in Fig. \ref{fig:interpretability-aerobearing}.

As marked in Fig. \ref{fig:interpretability-aerobearing}(a),  the frequency amplitude mainly exists in five information frequency bands that contain most information of aerospace bearing dataset, i.e., band 1-2-3-4-5. As shown in Fig. \ref{fig:interpretability-Houde}(b)-(c), the results are consistent with the interpretability analysis of the previous two datasets. Both two models pay correct attention to the information frequency bands of the aerospace bearing dataset, but the O-FR of TFN-STTF has one amplitude peak (nearly 0.35 in frequency) outside the information frequency bands, while  the O-FR of TFN-Chirplet has a very small amplitude peak (nearly 0.35 in frequency) outside the information frequency bands due to its capability to adjust filtering bandwidth. In conclusion, the O-FR of trained TFN-Chirplet has a good correspondence with the frequency spectrum of aerospace bearing dataset, demonstrating the outstanding interpretability of TFN-Chirplet again.

\section{Discussion}

To further analyze the properties of TFN,  the comparison with contrast models, the analysis of the training process and training time, the few-shot analysis , and the generalizability analysis are conducted on the CWRU open-source dataset.  The comparison part shows the essential differences between TFN and contrast models and explains the reason for the superior diagnostic performance of TFNs. The analysis of the training process and training time discusses the advantage of TFN in convergence speed in the training process and the time cost of TFN. The few-shot analysis explores the relationship between the diagnostic accuracy of TFN and the number of training samples, demonstrating the excellent few-shot ability of TFN.  The generalizability analysis proves the feasibility of generalizing the TFconv layer to other CNN models.

\subsection{Comparison with Contrast Models}
The motivation of contrast models (i.e., SincNet\cite{ravanelliInterpretableConvolutionalFilters2019a}, WKN\cite{liWaveletKernelNetInterpretableDeep2022}, and W-CNN\cite{gangulyWaveletKernelBased2020}) shown in Table \ref{tab:comparison with our work} is to parameterize or initialize traditional convolutional kernel by a specific kernel function, but these models only consider real value convolutional kernels and are equivalent to a series of bandpass FIR filters, whose outputs are filtered sub-signals. On the contrary, the TFconv layer uses Real-Imaginary Mechanism to simulate complex value convolutional kernels, thus the output of the TFconv layer is time-frequency distribution (the energy distribution of a signal in the time-frequency domain), which is the same as that of traditional time-frequency transform methods, except that TFconv layer is trainable and adaptive based on the specific dataset.

We demonstrate the differences between TFNs and contrast models in two ways: formula derivation and the output presentation of these models. As for the formula derivation part, we would show the difference between the output of "complex kernel" (used by TFNs) and that of "real kernel only" (used by contrast models). We take STFT, a basic time-frequency transform method, for example. Given an input signal $x(t)$, the process of "complex kernel" could be denoted as
\begin{equation}
X(\tau,\omega) = \int x(t) w(t-\tau)e^{-i\omega t} \ {\rm d} t
\label{eq:CK_process}
\end{equation}
where, $X(\tau,\omega)$ is the output of the process of "complex kernel", $\tau$ and $\omega$ are time shift parameter and frequency shift parameter, respectively. $w(t)$ is the window function of STFT, whose length is denoted as $T$. The process of "real kernel only" could be denoted as
\begin{equation}
\widehat{X(\tau,\omega)} = \int x(t) w(t-\tau) \cos{\omega t} \ {\rm d} t
\label{eq:RK_process}
\end{equation}
{}
According to Fourier expansion, $x(t) w(t-\tau)$ could be expanded as
\begin{equation}
x(t) w(t-\tau) = \frac{a_0(\tau)}{2} + \sum_{n=1}^{\infty}a_n(\tau) \sin{[n\omega_0t + \phi_n(\tau)]}
\label{eq:windowfunc_FE}
\end{equation}
where, $a_n(\tau)$ denotes the amplitude of the input signal $x(t)$ in time $\tau$ and frequency $n\omega_0$, and $\omega_0 = 2\pi /\ T$ is the base frequency.

Therefore, bringing Eq.\ref{eq:windowfunc_FE} back to Eq.\ref{eq:CK_process} with modulus operator and considering the orthogonality of the sine function, we could get
\begin{equation}
\begin{aligned}
|X(\tau, \omega)| & = \left| \int x(t) w(t-\tau)e^{-i\omega t} \ {\rm d} t \right| \\
& = \left| \int\left\{\frac{a_{0}(\tau)}{2}+\sum_{n=1}^{\infty} a_{n}(\tau) \sin \left[n \omega_{0} t+\phi_{n}(\tau)\right]\right\} e^{-i \omega t} \ {\rm d} t \right| \\
& =\left|\frac{T}{2} \cdot\left[a_{\omega}(\tau) \sin \left(\phi_{\omega}(\tau)\right)+a_{\omega}(\tau) \cos \left(\phi_{\omega}(\tau)\right) \cdot i\right]\right|\\
& =\frac{T}{2} a_{\omega}(\tau)
\end{aligned}
\label{eq:CK_process_simp}
\end{equation}
Above is the output of the process of "complex kernel", and $a_\omega (\tau)$ is the amplitude of the input signal x(t) in time $\tau$ and frequency $\omega$, i.e., the time-frequency distribution.

To get the output of the process of "real kernel only", it is the same to bring Eq.\ref{eq:windowfunc_FE} back to Eq.\ref{eq:RK_process} and we will get
\begin{equation}
\begin{aligned}
X \widehat{(\tau, \omega)} & =\int x(t) w(t-\tau) \cos (\omega t) d t \\
& =\int\left\{\frac{a_{0}(\tau)}{2}+\sum_{n=1}^{\infty} a_{n}(\tau) \sin \left[n \omega_{0} t+\phi_{n}(\tau)\right]\right\} \cos (\omega t) d t \\
& =\frac{T}{2} a_{\omega}(\tau) \cdot \sin \phi_{\omega}(\tau)
\end{aligned}
\label{eq:RK_process_simp}
\end{equation}
The output of "real kernel only" has two parts: the former part $a_\omega (\tau)$ is time-frequency distribution and the later part is the phrase information subject to time $\tau$ and frequency $\omega$. This is the reason that the output of "real kernel only" is filtered sub-signals, not the time-frequency distribution as processed by "complex kernel".

\begin{figure}[htbp]
        \centering
        \includegraphics[width=17 cm]{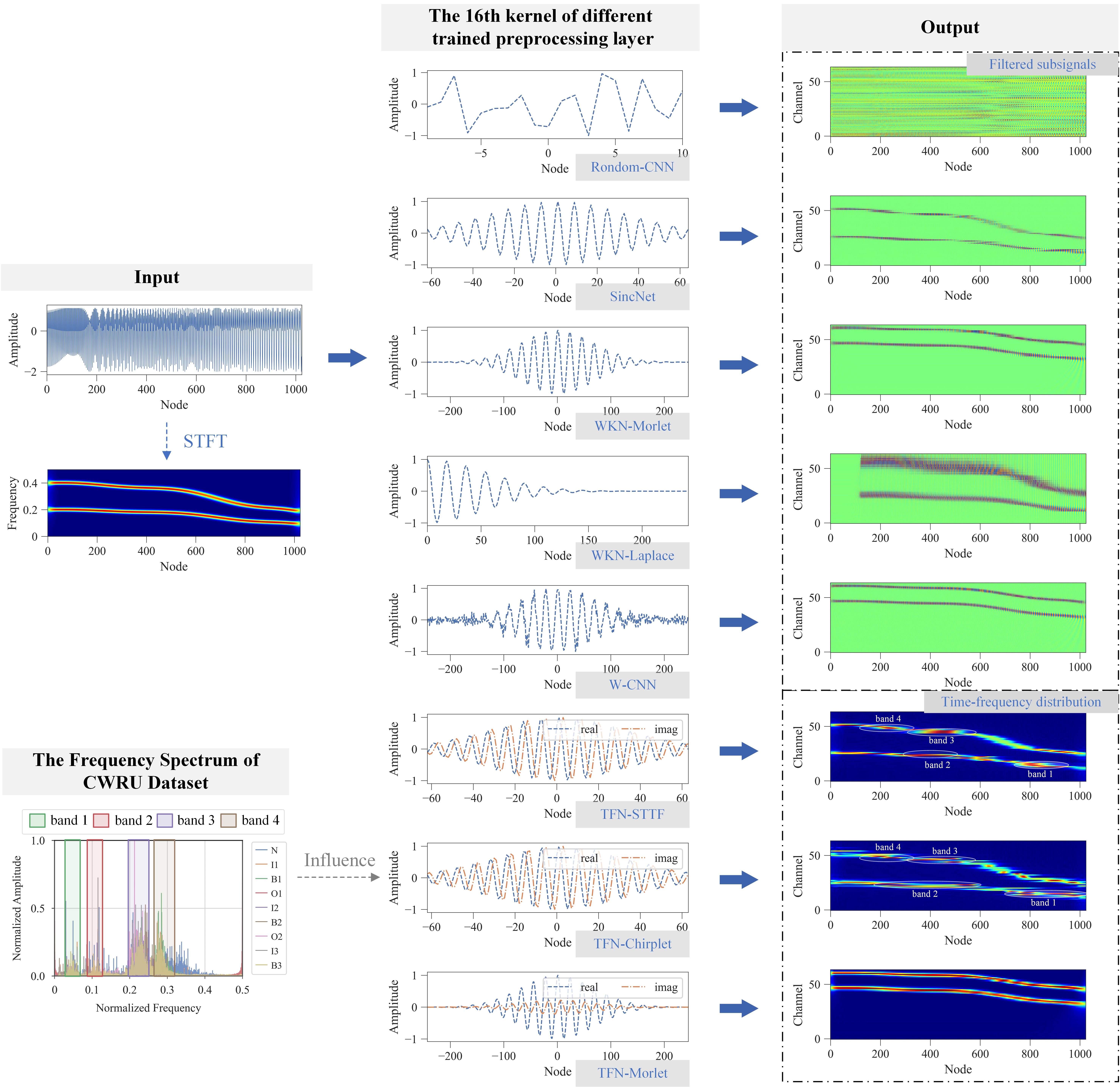}
        \vspace{-5 mm}
        \caption{The illustation of the processing process of different models.}
        \label{fig:difference}
\end{figure}

After the formula derivation part, we use the output presentation to further introduce the differences between TFNs and contrast models (i.e., SincNet, WKN-Morlet, WKN-Laplace, W-CNN). For all the models, the channel numbers of the preprocessing layer are set to 64, and the experiment settings are the same as the interpretability analysis on the CWRU dataset. The illustration of the processing process of different models is shown in Fig. \ref{fig:difference}, and the simulated input signal with its time-frequency distribution (processed by STFT) is shown in the left column of the figure. The 16th kernel of different trained preprocessing layers are shown in the middle column to show the model differences in kernel shape. The preprocessing layer outputs of different trained models are shown in the right column to show the model differences in output.

As we can see from the middle column, the kernel of Rondom-CNN is randomly initialized which is the way taht the traditional convolutional layer goes, and the contrast models use real value kernels and parameterize or initialize them to a specific frequency which is equivalent to a FIR filter, while TFNs (i.e., TFN-STTF, TFN-Chirplet, and TFN-Morlet) use complex value kernels and parameterize them, which is equivalent to time-frequency transform.

As we can see from the right column, the outputs of Rondom-CNN is hard to catch any useful information. With kernel functions embedded, the outputs of contrast models have a clear correspondence to the time-frequency distribution of the input signal, meaning these models can extract time-frequency information to some extent. However, these models only consider real value kernels and their outputs contain the phase information of the input signal as explained in Eq.\ref{eq:RK_process_simp}, making their ouputs blurry and still obviously different from the time-frequency distribution of the input signal processed by STFT. On the contrary, TFNs use Real-Imaginary Mechanism to take complex value kernels into consideration, and the outputs of TFNs are exactly corresponding to the time-frequency distribution of the input signal, except for some distortions on some specific frequency bands. It is easy to conclude that these distortions are caused by the training process of the kernel functions. As discussed in interpretability analysis (Fig.\ref{fig:interpretability}), TFNs will change their frequency response (FR) during the training process, and these specific frequency bands within distortions of TFN-STTF and TFN-Chirplet are  exactly corresponding to the information frequency bands shown in the frequency spectrum of CWRU dataset (left bottom in Fig. \ref{fig:difference}), which is also consistent with the phenomenon we discussed in the interpretability analysis of CWRU dataset before.

In conclusion, the outputs of contrast models are a series of filtered sub-signals, that have a certain correspondence to the time-frequency distribution of the input signal but still have distinct differences. The outputs of TFNs are the time-frequency distribution of the input signal, but with more focus on the information bands of the training dataset, and this explains the superior diagnostic performance of TFNs that fault-related features are extracted by the TFconv layer to facilitate the following fault classification.

\subsection{Training Process and Traing Time}
To further analyze our TFNs models, we compare the convergence speed of different models on the CWRU dataset. Backbone models (i.e., Backbone-CNN and Rondom-CNN), contrast models (i.e., SincNet, WKN-Morlet, WKN-Laplace, and W-CNN) and TFNs (i.e., TFN-STTF, TFN-Chirplet, and TFN-Morlet) are chosen for comparison. The channel numbers of the preprocessing layer are all set to 64, the training epoch is set to 80 to get complete records, and the rest experimental settings are the same as the diagnostic experiment on the CWRU dataset.

\begin{figure}[H]
        \centering
        \includegraphics[width=17 cm]{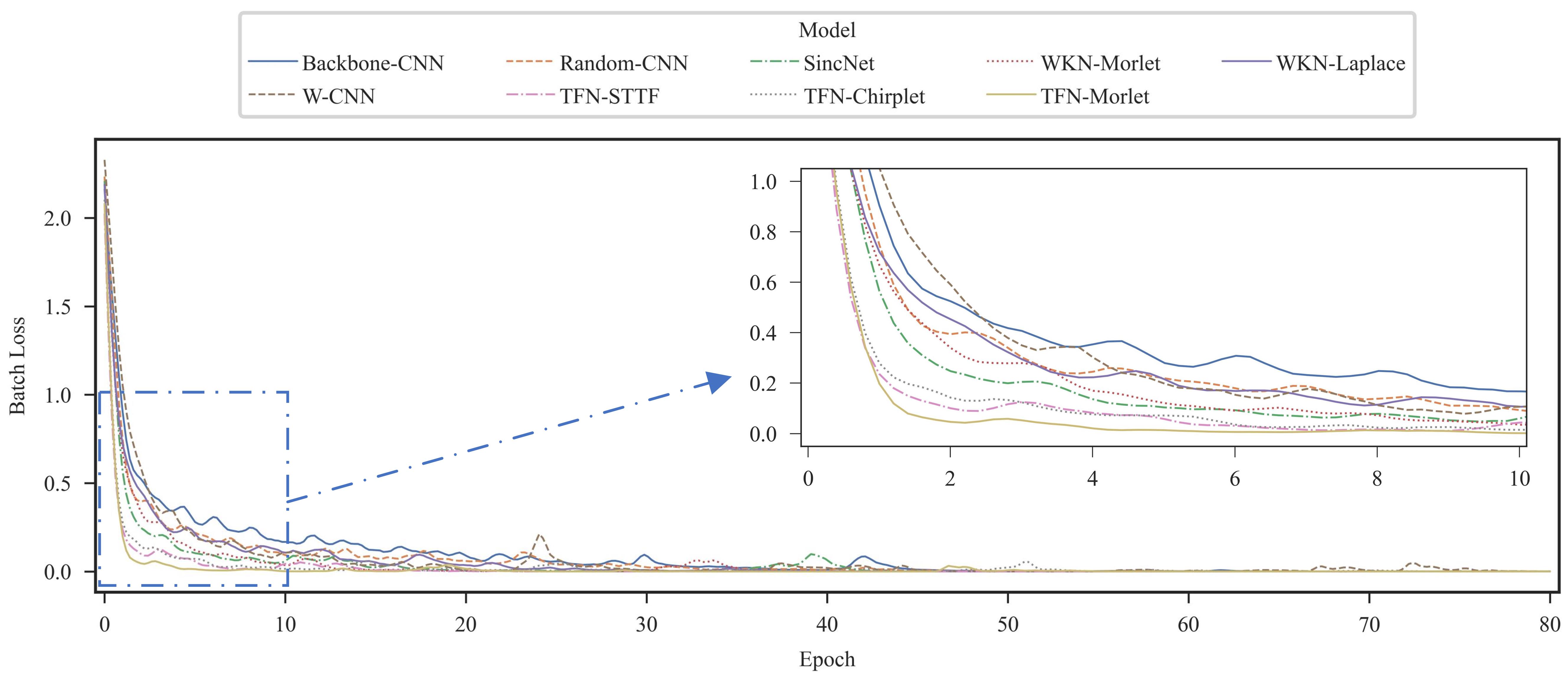}
        \vspace{-8 mm}
        \caption{The training process of different models on the CWRU dataset.}
        \label{fig:train_process}
\end{figure}

The training process of different models is shown in Fig.\ref{fig:train_process}, and Backbone-CNN gets the worst performance. Random-CNN, W-CNN, and WKN-Laplace perform much better than Backbone-CNN. SincNet and WKN-Morlet belong to the second tier, and they perform slightly better than the  previous three models. TFNs have a fast convergence speed than all other models, and TFN-Morlet performs best among them. The convergence speed is consistent with the diagnostic performance shown in Fig.\ref{fig:CWRU acc}, and the fast convergence speed of TFNs is attributed to their ability to extract time-frequency features.

\begin{figure}[htbp]
        \centering
        \includegraphics[width=17 cm]{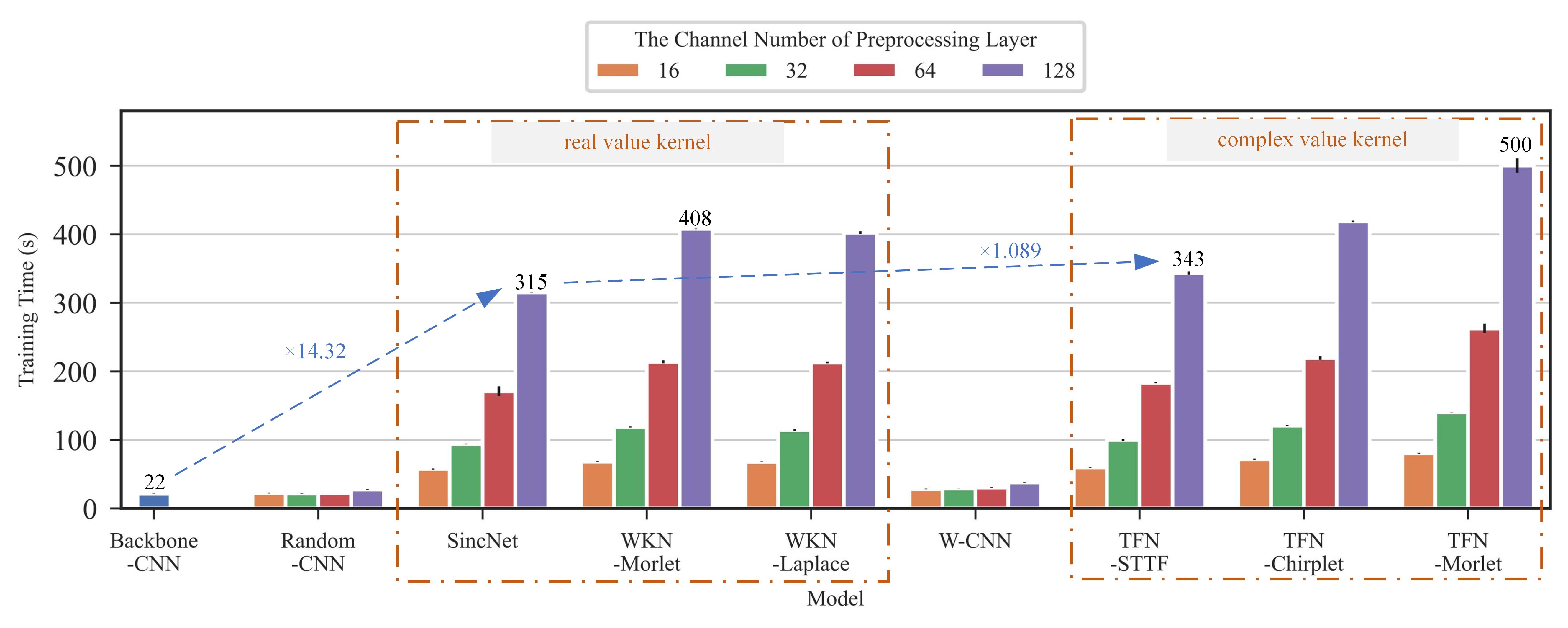}
        \vspace{-8 mm}
        \caption{The training time of different models with different preprocessing layer channels on the CWRU dataset.}
        \label{fig:train_time}
\end{figure}

Parameterizing convolutional kernels can significantly improve the diagnostic ability of CNN models, but the diagnostic improvement is at the cost of training time, which  is barely mentioned in the current literature \cite{liWaveletKernelNetInterpretableDeep2022,ravanelliInterpretableConvolutionalFilters2019a,gangulyWaveletKernelBased2020}. To quantify the time cost of parameterizing convolutional kernel, we recorded the training time of backbone models, contrast models and TFNs with different channel numbers in the diagnostic experiment on CWRU dataset, which is shown in Fig.\ref{fig:train_time}.

\begin{enumerate}
        \item The training time of Backbone-CNN and Rondom-CNN is nearly 22 seconds per training. W-CNN just initializes the convolutional kernel, not parameterizes, so the training time of W-CNN is nearly 30 seconds per training, which is close to Backbone models.
        \item SincNet, WKN-Morlet and WKN-Laplace parameterize the real value convolutional kernels, so their training times increase significantly than Backbone-CNN and that with 128 channels are close to 350 seconds, which is almost 16 times more than Backbone models. Within these three models, the training times of WKN-Morlet and WKN-Laplace are more than SincNet due to their longer kernel length shown in Table. \ref{tab:kernel Function}.
        \item TFN-STTF, TFN-Chirplet and TFN-Morlet parameterize the complex value convolutional kernels, and their training times with 128 channels are close to 400 seconds, which is almost 18 times more than Backbone models. Within these three models, TFN-Chirplet requires more training time than TFN-STTF because Chirplet kernel function has an additional control parameter (i.e., the linear frequency modulation factor $\alpha$) to train, and TFN-Morlet requires the most training time because morlet kernel function has the longest kernel length than other functions.
        \item As the number of channels increases, the training time of these models that parameterize convolutional kernel increases significantly, so the channel number of such models needs to be carefully considered to strike a balance between diagnostic accuracy and training time.
\end{enumerate} 

In conclusion, parameterizing convolutional kernel is extremely time-consuming, and such models require numerous training time than backbone models, while parameterizing complex value kernel (used by TFNs) is only slightly more time-consuming than parameterizing real value kernel (used by contrast models\cite{liWaveletKernelNetInterpretableDeep2022,ravanelliInterpretableConvolutionalFilters2019a}), which makes our TFNs still competitive compared to contrast models.

\subsection{Few-shot Analysis}
To further analyze the few-shot ability of our TFN models, we conduct a few-shot experiment on CWRU dataset. The channel numbers of the preprocessing layer are all set to 64, and the experimental settings are the same as the diagnostic performance experiment on CWRU dataset where each class has 450 samples, for a total of 4500 samples. To test the few-shot ability of TFNs, we pick a certain number (i.e., 5, 10, 20, 50, 100, 150, 200, 250 and 300) of samples each class as training data, and the remaining samples are used for testing. Considering the difference in the number of training samples, the number of training epochs changes accordingly to ensure that the number of training batch is as equal as possible but no more than 300, which is designed as follows
\begin{equation}
N_{\rm epoch} = \min \left(\frac{50 \times 300}{N_{\rm training\_sample}}, 300\right)
\label{eq:few-shot}
\end{equation}
where, $N_{\rm epoch}$ denotes the number of training epochs and $N_{\rm training\_sample}$ denotes the number of training samples each class.

\begin{figure}[htbp]
        \centering
        \includegraphics[width=17 cm]{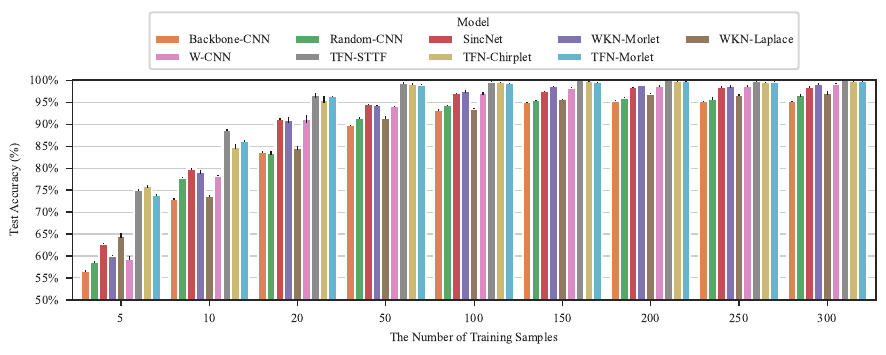}
        \vspace{-8 mm}
        \caption{The diagnostic accuracy of different models with different numbers of training samples on the CWRU dataset.}
        \label{fig:few-shot}
\end{figure}

The result of the few-shot experiment is shown in Fig. \ref{fig:few-shot}. When the number of training samples is 5, the diagnostic accuracies of the models are all below 80\%. But our TFNs (close to 75\%) perform more outstandingly than other models (below 65\%), and the gap in diagnostic accuracy exceeds 10\%. When the number of training samples increases to 50, the diagnostic accuracy of TFNs is close to 100\%, and the diagnostic accuracy gap between TFNs and other models has been narrowed to about 5\%. After that, with the increase in the number of training samples, the performance of TFNs remains at about 100\%, and the diagnostic accuracy of other models gradually increases, and finally, the accuracy gap is about 1$ \sim $2\%. In conclusion, our proposed TFNs have much better performances than other models in few-shot diagnostic tasks, and this outstanding performance is due to their ability to extract time-frequency features demonstrated in Fig.\ref{fig:train_process}.

\subsection{Generalizability}

In order to verify the generalizability of the TFconv layer, three typical CNNs with different depths are selected as the backbone to obtain different TFNs. The vibration signal under 3 HP load condition of the CWRU dataset is used as the input samples, and the channel numbers of the TFconv layer are all set to 128. Other experimental settings are consistent with those in the previous diagnostic experiments on the CWRU dataset, and the diagnosis results are shown in Table \ref{tab:generalization}.

\begin{table}[htbp]
\caption{The diagnostic Results of Generalizing TFconv Layers to Other CNN Architectures \label{tab:generalization}}
\centering
\begin{tabular}{lllll}
\toprule[1pt]
Backbone & Combined with & Accuracy & Variance\\
\midrule[0.5pt]
LeNet   & None            & 90.02          & 0.177\\
        & TFconv-STTF     & 95.71          & 0.355\\
        & TFconv-Chirplet & 94.25          & 0.339\\
        & TFconv-Morlet   & \textbf{98.96} & 0.346\\
\midrule[0.5pt]
AlexNet & None            & 97.32          & 0.476\\
        & TFconv-STTF     & 98.27          & 0.413\\
        & TFconv-Chirplet & 97.84          & 0.339\\
        & TFconv-Morlet   & \textbf{99.89 }& 0.129\\
\midrule[0.5pt]
ResNet  & None            & 97.74          & 0.243\\
        & TFconv-STTF     & 99.58          & 0.183\\
        & TFconv-Chirplet & 98.79          & 0.464\\
        & TFconv-Morlet   & \textbf{99.96} & 0.058\\
\bottomrule[1pt]
\end{tabular}
\end{table}

It can be seen from the results that, the TFconv layer can significantly improve the diagnostic accuracy on the basis of different backbone CNNs. The TFconv layer with the Morlet kernel always achieves the best diagnostic performance, followed by the STTF kernel. The Chirplet kernel is slightly worse than the Morlet kernel and STTF kernel, but still better than the backbone model. This experiment shows that the proposed TFconv layer is a general method and can be applied to other CNNs with different depths to improve their diagnostic performance. Besides, this experiment illustrates the importance of the backbone model. We recommend researchers adopt the TFconv layer with STTF or Morlet kernel functions, and then combine a backbone CNN with enough depth to formulate the TFN.

\section{Conclusion}

In this article, an interpretable time-frequency convolutional (TFconv) layer is proposed to extract fault-related time-frequency information. Taking the TFconv layer as a preprocessing layer, we formulated the Time-Frequency Network (TFN) to achieve higher diagnostic accuracy and explain the focusing frequency area in the prediction-making of CNN models. The diagnostic effectiveness and interpretability of TFN have been verified by three sets of mechanical fault diagnosis experiments. The conclusions of this article could be summarized as follows: 1) The participation of the TFconv layer can greatly improve the diagnostic performance of the CNN in mechanical fault diagnosis tasks. 2) The TFconv layer could explain the focusing frequency area of TFN to extract features and make predictions. 3) The kernel function and channel number of the TFconv layer have a great influence on the diagnostic performance of TFN, and the TFconv layer with 64-channel STTF kernel can achieve the overall optimum in accuracy and efficiency. 4) The TFconv layer has outstanding performance on convergence speed and few-shot scenarios, and can be generalized to other CNN models with different depths. In future research, we will explore other kernel functions with adaptive frequency bandwidth to interpret focusing frequency more explicitly, and investigate the effectiveness of the TFconv layer for other neural networks other than CNNs (e.g. autoencoder).

\section*{Acknowledge}
This project was supported  by the National Natural Science Foundation of China under Grant No. 12272219, and No. 12121002.

\normalem
\footnotesize
\bibliographystyle{elsarticle-num}
\bibliography{reference.bib}

\begin{thebibliography}{10}
\expandafter\ifx\csname url\endcsname\relax
  \def\url#1{\texttt{#1}}\fi
\expandafter\ifx\csname urlprefix\endcsname\relax\def\urlprefix{URL }\fi
\expandafter\ifx\csname href\endcsname\relax
  \def\href#1#2{#2} \def\path#1{#1}\fi

\bibitem{liRollingBearingFault2021}
J.~Li, X.~Wang, H.~Wu, Rolling {{Bearing Fault Detection Based}} on {{Improved
  Piecewise Unsaturated Bistable Stochastic Resonance Method}}, IEEE
  Transactions on Instrumentation and Measurement 70 (2021) 1--9.
\newblock \href {https://doi.org/10.1109/TIM.2020.3024038}
  {\path{doi:10.1109/TIM.2020.3024038}}.

\bibitem{LI2015330}
Y.~Li, M.~Xu, Y.~Wei, W.~Huang, An improvement {{EMD}} method based on the
  optimized rational {{Hermite}} interpolation approach and its application to
  gear fault diagnosis, Measurement 63 (2015) 330--345.
\newblock \href {https://doi.org/10.1016/j.measurement.2014.12.021}
  {\path{doi:10.1016/j.measurement.2014.12.021}}.

\bibitem{taoDatadrivenSmartManufacturing2018}
F.~Tao, Q.~Qi, A.~Liu, A.~Kusiak, Data-driven smart manufacturing, Journal of
  Manufacturing Systems 48 (2018) 157--169.
\newblock \href {https://doi.org/10.1016/j.jmsy.2018.01.006}
  {\path{doi:10.1016/j.jmsy.2018.01.006}}.

\bibitem{chenDataDrivenFaultDiagnosis2022}
H.~Chen, B.~Jiang, S.~X. Ding, B.~Huang, Data-{{Driven Fault Diagnosis}} for
  {{Traction Systems}} in {{High-Speed Trains}}: {{A Survey}}, {{Challenges}},
  and {{Perspectives}}, IEEE Transactions on Intelligent Transportation Systems
  23~(3) (2022) 1700--1716.
\newblock \href {https://doi.org/10.1109/TITS.2020.3029946}
  {\path{doi:10.1109/TITS.2020.3029946}}.

\bibitem{leiApplicationsMachineLearning2020}
Y.~Lei, B.~Yang, X.~Jiang, F.~Jia, N.~Li, A.~K. Nandi, Applications of machine
  learning to machine fault diagnosis: {{A}} review and roadmap, Mechanical
  Systems and Signal Processing 138 (2020).
\newblock \href {https://doi.org/10.1016/j.ymssp.2019.106587}
  {\path{doi:10.1016/j.ymssp.2019.106587}}.

\bibitem{liRotationalMachineHealth2012}
R.~Li, D.~He, Rotational {{Machine Health Monitoring}} and {{Fault Detection
  Using EMD-Based Acoustic Emission Feature Quantification}}, IEEE Transactions
  on Instrumentation and Measurement 61~(4) (2012) 990--1001.
\newblock \href {https://doi.org/10.1109/TIM.2011.2179819}
  {\path{doi:10.1109/TIM.2011.2179819}}.

\bibitem{siKeyPerformanceIndicatorRelatedProcessMonitoring2021}
Y.~Si, Y.~Wang, D.~Zhou, Key-{{Performance-Indicator-Related Process Monitoring
  Based}} on {{Improved Kernel Partial Least Squares}}, IEEE Transactions on
  Industrial Electronics 68~(3) (2021) 2626--2636.
\newblock \href {https://doi.org/10.1109/TIE.2020.2972472}
  {\path{doi:10.1109/TIE.2020.2972472}}.

\bibitem{pengApplicationWaveletTransform2004}
Z.~Peng, F.~Chu, Application of the wavelet transform in machine condition
  monitoring and fault diagnostics: A review with bibliography, Mechanical
  Systems and Signal Processing 18~(2) (2004) 199--221.
\newblock \href {https://doi.org/10.1016/S0888-3270(03)00075-X}
  {\path{doi:10.1016/S0888-3270(03)00075-X}}.

\bibitem{liApplicationBandwidthEMD2017}
Y.~Li, M.~Xu, X.~Liang, W.~Huang, Application of {{Bandwidth EMD}} and
  {{Adaptive Multiscale Morphology Analysis}} for {{Incipient Fault Diagnosis}}
  of {{Rolling Bearings}}, IEEE Transactions on Industrial Electronics 64~(8)
  (2017) 6506--6517.
\newblock \href {https://doi.org/10.1109/TIE.2017.2650873}
  {\path{doi:10.1109/TIE.2017.2650873}}.

\bibitem{brunettiComputerVisionDeep2018}
A.~Brunetti, D.~Buongiorno, G.~F. Trotta, V.~Bevilacqua, Computer vision and
  deep learning techniques for pedestrian detection and tracking: {{A}} survey,
  Neurocomputing 300 (2018) 17--33.
\newblock \href {https://doi.org/10.1016/j.neucom.2018.01.092}
  {\path{doi:10.1016/j.neucom.2018.01.092}}.

\bibitem{fayekEvaluatingDeepLearning2017}
H.~M. Fayek, M.~Lech, L.~Cavedon, Evaluating deep learning architectures for
  {{Speech Emotion Recognition}}, Neural Networks 92 (2017) 60--68.
\newblock \href {https://doi.org/10.1016/j.neunet.2017.02.013}
  {\path{doi:10.1016/j.neunet.2017.02.013}}.

\bibitem{silverMasteringGameGo2017}
D.~Silver, J.~Schrittwieser, K.~Simonyan, I.~Antonoglou, A.~Huang, A.~Guez,
  T.~Hubert, L.~Baker, M.~Lai, A.~Bolton, Y.~Chen, T.~Lillicrap, F.~Hui,
  L.~Sifre, G.~{van den Driessche}, T.~Graepel, D.~Hassabis, Mastering the game
  of {{Go}} without human knowledge, Nature 550~(7676) (2017) 354--359.
\newblock \href {https://doi.org/10.1038/nature24270}
  {\path{doi:10.1038/nature24270}}.

\bibitem{shaoElectricLocomotiveBearing2018}
H.~Shao, H.~Jiang, H.~Zhang, T.~Liang, Electric {{Locomotive Bearing Fault
  Diagnosis Using}} a {{Novel Convolutional Deep Belief Network}}, IEEE
  Transactions on Industrial Electronics 65~(3) (2018) 2727--2736.
\newblock \href {https://doi.org/10.1109/TIE.2017.2745473}
  {\path{doi:10.1109/TIE.2017.2745473}}.

\bibitem{wangCoarsetoFineProgressiveKnowledge2021}
Y.~Wang, R.~Liu, D.~Lin, D.~Chen, P.~Li, Q.~Hu, C.~L.~P. Chen,
  Coarse-to-{{Fine}}: {{Progressive Knowledge Transfer-Based Multitask
  Convolutional Neural Network}} for {{Intelligent Large-Scale Fault
  Diagnosis}}, IEEE Transactions on Neural Networks and Learning Systems
  (2021).
\newblock \href {https://doi.org/10.1109/TNNLS.2021.3100928}
  {\path{doi:10.1109/TNNLS.2021.3100928}}.

\bibitem{zhaoIntelligentFaultDiagnosis2022}
X.~Zhao, J.~Yao, W.~Deng, P.~Ding, Y.~Ding, M.~Jia, Z.~Liu, Intelligent {{Fault
  Diagnosis}} of {{Gearbox Under Variable Working Conditions With Adaptive
  Intraclass}} and {{Interclass Convolutional Neural Network}}, IEEE
  Transactions on Neural Networks and Learning Systems (2022) 1--15\href
  {https://doi.org/10.1109/TNNLS.2021.3135877}
  {\path{doi:10.1109/TNNLS.2021.3135877}}.

\bibitem{pengMultibranchMultiscaleCNN2020}
D.~Peng, H.~Wang, Z.~Liu, W.~Zhang, M.~J. Zuo, J.~Chen, Multibranch and
  {{Multiscale CNN}} for {{Fault Diagnosis}} of {{Wheelset Bearings Under
  Strong Noise}} and {{Variable Load Condition}}, IEEE Transactions on
  Industrial Informatics 16~(7) (2020) 4949--4960.
\newblock \href {https://doi.org/10.1109/TII.2020.2967557}
  {\path{doi:10.1109/TII.2020.2967557}}.

\bibitem{nieNovelNormalizedRecurrent2021}
X.~Nie, G.~Xie, A novel normalized recurrent neural network for fault diagnosis
  with noisy labels, Journal of Intelligent Manufacturing 32~(5) (2021)
  1271--1288.
\newblock \href {https://doi.org/10.1007/s10845-020-01608-8}
  {\path{doi:10.1007/s10845-020-01608-8}}.

\bibitem{zhaoDeepLearningAlgorithms2020}
Z.~Zhao, T.~Li, J.~Wu, C.~Sun, S.~Wang, R.~Yan, X.~Chen, Deep learning
  algorithms for rotating machinery intelligent diagnosis: {{An}} open source
  benchmark study, ISA Transactions 107 (2020) 224--255.
\newblock \href {https://doi.org/10.1016/j.isatra.2020.08.010}
  {\path{doi:10.1016/j.isatra.2020.08.010}}.

\bibitem{zhangVisualInterpretabilityDeep2018}
Q.-s. Zhang, S.-c. Zhu, Visual interpretability for deep learning: A survey,
  Frontiers of Information Technology \& Electronic Engineering 19~(1) (2018)
  27--39.
\newblock \href {https://doi.org/10.1631/FITEE.1700808}
  {\path{doi:10.1631/FITEE.1700808}}.

\bibitem{xiLeastSquaresSupport2019}
P.-P. Xi, Y.-P. Zhao, P.-X. Wang, Z.-Q. Li, Y.-T. Pan, F.-Q. Song, Least
  squares support vector machine for class imbalance learning and their
  applications to fault detection of aircraft engine, Aerospace Science and
  Technology 84 (2019) 56--74.
\newblock \href {https://doi.org/10.1016/j.ast.2018.08.042}
  {\path{doi:10.1016/j.ast.2018.08.042}}.

\bibitem{ivanovsPerturbationbasedMethodsExplaining2021}
M.~Ivanovs, R.~Kadikis, K.~Ozols, Perturbation-based methods for explaining
  deep neural networks: {{A}} survey, Pattern Recognition Letters 150 (2021)
  228--234.
\newblock \href {https://doi.org/10.1016/j.patrec.2021.06.030}
  {\path{doi:10.1016/j.patrec.2021.06.030}}.

\bibitem{fanInterpretabilityArtificialNeural2021}
F.-L. Fan, J.~Xiong, M.~Li, G.~Wang, On {{Interpretability}} of {{Artificial
  Neural Networks}}: {{A Survey}}, IEEE Transactions on Radiation and Plasma
  Medical Sciences 5~(6) (2021) 741--760.
\newblock \href {https://doi.org/10.1109/TRPMS.2021.3066428}
  {\path{doi:10.1109/TRPMS.2021.3066428}}.

\bibitem{zhangSurveyNeuralNetwork2021}
Y.~Zhang, P.~Tino, A.~Leonardis, K.~Tang, A {{Survey}} on {{Neural Network
  Interpretability}}, IEEE Transactions on Emerging Topics in Computational
  Intelligence 5~(5) (2021) 726--742.
\newblock \href {https://doi.org/10.1109/TETCI.2021.3100641}
  {\path{doi:10.1109/TETCI.2021.3100641}}.

\bibitem{dhurandharExplanationsBasedMissing2018}
A.~Dhurandhar, P.-Y. Chen, R.~Luss, C.-C. Tu, P.~Ting, K.~Shanmugam, P.~Das,
  Explanations based on the {{Missing}}: {{Towards Contrastive Explanations}}
  with {{Pertinent Negatives}}, in: Proc. {{Adv}}. {{Neural Inf}}. {{Process}}.
  {{Syst}}. ({{NeurIPS}}), Vol.~31, {Curran Associates, Inc.}, 2018.

\bibitem{wangInterpretNeuralNetworks2018}
Y.~Wang, H.~Su, B.~Zhang, X.~Hu, Interpret {{Neural Networks}} by {{Identifying
  Critical Data Routing Paths}}, in: Proc. {{IEEE Conf}}. {{Comput}}. {{Vis}}.
  {{Pattern Recog}}. ({{CVPR}}), {IEEE}, {Salt Lake City, UT}, 2018, pp.
  8906--8914.
\newblock \href {https://doi.org/10.1109/CVPR.2018.00928}
  {\path{doi:10.1109/CVPR.2018.00928}}.

\bibitem{bauNetworkDissectionQuantifying2017}
D.~Bau, B.~Zhou, A.~Khosla, A.~Oliva, A.~Torralba, Network {{Dissection}}:
  {{Quantifying Interpretability}} of {{Deep Visual Representations}}, in:
  Proc. {{IEEE Conf}}. {{Comput}}. {{Vis}}. {{Pattern Recog}}. ({{CVPR}}),
  {IEEE}, {Honolulu, HI}, 2017, pp. 3319--3327.
\newblock \href {https://doi.org/10.1109/CVPR.2017.354}
  {\path{doi:10.1109/CVPR.2017.354}}.

\bibitem{simonyanDeepConvolutionalNetworks2014}
K.~Simonyan, A.~Vedaldi, A.~Zisserman, Deep {{Inside Convolutional Networks}}:
  {{Visualising Image Classification Models}} and {{Saliency Maps}} (Apr.
  2014).
\newblock \href {http://arxiv.org/abs/1312.6034} {\path{arXiv:1312.6034}}.

\bibitem{selvarajuGradCAMVisualExplanations2017}
R.~R. Selvaraju, M.~Cogswell, A.~Das, R.~Vedantam, D.~Parikh, D.~Batra,
  Grad-{{CAM}}: {{Visual Explanations}} from {{Deep Networks}} via
  {{Gradient-Based Localization}}, in: Proc. {{IEEE Int}}. {{Conf}}.
  {{Comput}}. {{Vis}}. ({{ICCV}}), {IEEE}, {Venice}, 2017, pp. 618--626.
\newblock \href {https://doi.org/10.1109/ICCV.2017.74}
  {\path{doi:10.1109/ICCV.2017.74}}.

\bibitem{sundararajanAxiomaticAttributionDeep2017}
M.~Sundararajan, A.~Taly, Q.~Yan, Axiomatic {{Attribution}} for {{Deep
  Networks}}, in: D.~Precup, Y.~W. Teh (Eds.), Proceedings of the 34th
  {{International Conference}} on {{Machine Learning}}, Vol.~70 of Proceedings
  of {{Machine Learning Research}}, {PMLR}, 2017, pp. 3319--3328.

\bibitem{kohUnderstandingBlackboxPredictions2017}
P.~W. Koh, P.~Liang, Understanding {{Black-box Predictions}} via {{Influence
  Functions}}, in: D.~Precup, Y.~W. Teh (Eds.), Proceedings of the 34th
  {{International Conference}} on {{Machine Learning}}, Vol.~70 of Proceedings
  of {{Machine Learning Research}}, {PMLR}, 2017, pp. 1885--1894.

\bibitem{liWhiteningNetGeneralizedNetwork2021}
J.~Li, Y.~Wang, Y.~Zi, Z.~Zhang, Whitening-{{Net}}: {{A Generalized Network}}
  to {{Diagnose}} the {{Faults Among Different Machines}} and {{Conditions}},
  IEEE Transactions on Neural Networks and Learning Systems (2021) 1--14\href
  {https://doi.org/10.1109/TNNLS.2021.3071564}
  {\path{doi:10.1109/TNNLS.2021.3071564}}.

\bibitem{wuHybridClassificationAutoencoder2021}
X.~Wu, Y.~Zhang, C.~Cheng, Z.~Peng, A hybrid classification autoencoder for
  semi-supervised fault diagnosis in rotating machinery, Mechanical Systems and
  Signal Processing 149 (2021) 107327.
\newblock \href {https://doi.org/10.1016/j.ymssp.2020.107327}
  {\path{doi:10.1016/j.ymssp.2020.107327}}.

\bibitem{wangFullyInterpretableNeural2022}
D.~Wang, Y.~Chen, C.~Shen, J.~Zhong, Z.~Peng, C.~Li, Fully interpretable neural
  network for locating resonance frequency bands for machine condition
  monitoring, Mechanical Systems and Signal Processing 168 (2022) 108673.
\newblock \href {https://doi.org/10.1016/j.ymssp.2021.108673}
  {\path{doi:10.1016/j.ymssp.2021.108673}}.

\bibitem{zhaoInterpretableDenoisingLayer2021}
B.~Zhao, C.~Cheng, G.~Tu, Z.~Peng, Q.~He, G.~Meng, An {{Interpretable Denoising
  Layer}} for {{Neural Networks Based}} on {{Reproducing Kernel Hilbert Space}}
  and its {{Application}} in {{Machine Fault Diagnosis}}, Chinese Journal of
  Mechanical Engineering 34~(1) (2021) 44.
\newblock \href {https://doi.org/10.1186/s10033-021-00564-5}
  {\path{doi:10.1186/s10033-021-00564-5}}.

\bibitem{liWaveletKernelNetInterpretableDeep2022}
T.~Li, Z.~Zhao, C.~Sun, L.~Cheng, X.~Chen, R.~Yan, R.~X. Gao,
  {{WaveletKernelNet}}: {{An Interpretable Deep Neural Network}} for
  {{Industrial Intelligent Diagnosis}}, IEEE Transactions on Systems, Man, and
  Cybernetics: Systems 52~(4) (2022) 2302--2312.
\newblock \href {https://doi.org/10.1109/TSMC.2020.3048950}
  {\path{doi:10.1109/TSMC.2020.3048950}}.

\bibitem{ravanelliInterpretableConvolutionalFilters2019a}
M.~Ravanelli, Y.~Bengio, Interpretable {{Convolutional Filters}} with
  {{SincNet}} (Aug. 2019).
\newblock \href {http://arxiv.org/abs/1811.09725} {\path{arXiv:1811.09725}}.

\bibitem{gangulyWaveletKernelBased2020}
B.~Ganguly, S.~Chaudhury, S.~Biswas, D.~Dey, S.~Munshi, B.~Chatterjee,
  S.~Dalai, S.~Chakravorti, Wavelet {{Kernel}} based {{Convolutional Neural
  Network}} for {{Localization}} of {{Partial Discharge Sources}} within a
  {{Power Apparatus}}, IEEE Transactions on Industrial Informatics (2020)
  1--1\href {https://doi.org/10.1109/TII.2020.2991686}
  {\path{doi:10.1109/TII.2020.2991686}}.

\bibitem{michauFullyLearnableDeep2022}
G.~Michau, G.~Frusque, O.~Fink, Fully learnable deep wavelet transform for
  unsupervised monitoring of high-frequency time series, Proceedings of the
  National Academy of Sciences 119~(8) (2022) e2106598119.
\newblock \href {https://doi.org/10.1073/pnas.2106598119}
  {\path{doi:10.1073/pnas.2106598119}}.

\bibitem{yangMulticomponentSignalAnalysis2013}
Y.~Yang, W.~Zhang, Z.~Peng, G.~Meng, Multicomponent {{Signal Analysis Based}}
  on {{Polynomial Chirplet Transform}}, IEEE Transactions on Industrial
  Electronics 60~(9) (2013) 3948--3956.
\newblock \href {https://doi.org/10.1109/TIE.2012.2206331}
  {\path{doi:10.1109/TIE.2012.2206331}}.

\bibitem{tuIterativeNonlinearChirp2020}
G.~Tu, X.~Dong, S.~Chen, B.~Zhao, L.~Hu, Z.~Peng, Iterative nonlinear chirp
  mode decomposition: {{A Hilbert-Huang}} transform-like method in capturing
  intra-wave modulations of nonlinear responses, Journal of Sound and Vibration
  485 (2020) 115571.
\newblock \href {https://doi.org/10.1016/j.jsv.2020.115571}
  {\path{doi:10.1016/j.jsv.2020.115571}}.

\bibitem{cohenBetterWayDefine2019}
M.~X. Cohen, A better way to define and describe {{Morlet}} wavelets for
  time-frequency analysis, NeuroImage 199 (2019) 81--86.
\newblock \href {https://doi.org/10.1016/j.neuroimage.2019.05.048}
  {\path{doi:10.1016/j.neuroimage.2019.05.048}}.

\bibitem{andrearczykUsingFilterBanks2016}
V.~Andrearczyk, P.~F. Whelan, Using filter banks in convolutional neural
  networks for texture classification, Pattern Recognition Letters 84 (2016)
  63--69.

\bibitem{oppenheim1997signals}
A.~V. Oppenheim, A.~S. Willsky, S.~H. Nawab, G.~M. Hern{\'a}ndez, et~al.,
  Signals \& Systems, {Pearson Educaci\'on}, 1997.

\bibitem{smithRollingElementBearing2015}
W.~A. Smith, R.~B. Randall, Rolling element bearing diagnostics using the
  {{Case Western Reserve University}} data: {{A}} benchmark study, Mechanical
  Systems and Signal Processing 64--65 (2015) 100--131.
\newblock \href {https://doi.org/10.1016/j.ymssp.2015.04.021}
  {\path{doi:10.1016/j.ymssp.2015.04.021}}.

\end{thebibliography}

\end{document}